\documentclass[10pt,journal,compsoc]{IEEEtran}
\usepackage[nocompress]{cite}
\usepackage{algorithmic}
\usepackage{array}
\usepackage{mdwmath}
\usepackage{mdwtab}
\usepackage{eqparbox}
\usepackage[tight,footnotesize]{subfigure}
\usepackage{cite}
\usepackage{url}
\usepackage{multicol,multirow}
\usepackage[cmex10]{amsmath}
\usepackage{makeidx}
\usepackage{diagbox}
\usepackage{overpic}
\usepackage{rotating,soul}
\usepackage{wrapfig}
\usepackage{booktabs}
\usepackage[usenames,dvipsnames]{color}
\usepackage{ragged2e}
\usepackage{times}
\usepackage{graphicx}
\usepackage{amssymb}
\usepackage{tabularx}
\usepackage{xcolor}

\usepackage[normalem]{ulem}
\RequirePackage{silence}
\makeindex

\DeclareGraphicsExtensions{.pdf,.jpg,.png}

\usepackage[colorlinks,bookmarksnumbered]{hyperref}

\newcommand{\figref}[1]{Fig.~\ref{#1}}
\newcommand{\tabref}[1]{Tab.~\ref{#1}}
\newcommand{\equref}[1]{Eqn.~(\ref{#1})}
\newcommand{\secref}[1]{Sec.~\ref{#1}}

\newcommand{\myPara}[1]{\vspace{12pt} \noindent\textbf{#1} }
\newcommand{\conv}[2]{\texttt{conv#1\_#2}}

\DeclareMathOperator*{\argmax}{arg\,max}

\newcommand\chfrac[2]{\frac{\;\;\;\;#1\;\;\;\;}{\;\;\;\;#2\;\;\;\;}}

\def\ie{\emph{i.e.,~}}
\def\eg{\emph{e.g.,~}}
\def\etal{{\em et al.~}}
\def\etc{{\em etc.~}}
\def\wrt{{\em w.r.t.~}}

\def\Feature{\mathbf{F}}
\def\Activation{\mathbf{A}}
\def\Gradient{\mathbf{G}}

\def\wrt{\textit{w.r.t.}~}

\graphicspath{{./figure/}, {./figure/Related/}, {./figure/Author/}}

\begin{document}
	
\title{Deeply Explain CNN via Hierarchical Decomposition}

\author{Ming-Ming Cheng*, Peng-Tao Jiang*, Ling-Hao Han, Liang Wang, Philip Torr
  \IEEEcompsocitemizethanks{
    \IEEEcompsocthanksitem M.M. Cheng, P.T. Jiang, L.H. Han 
        are with TKLNDST, CS, Nankai University.
        M.M. Cheng is the corresponding author (cmm@nankai.edu.cn).
        * denotes equal contribution.
    \IEEEcompsocthanksitem L. Wang is with NLPR.
    \IEEEcompsocthanksitem P. Torr is with the University of Oxford.
    }
}

\markboth{IEEE TRANSACTIONS ON PATTERN ANALYSIS AND MACHINE INTELLIGENCE,~Vol.~xx, No. ~xx,~xxx.~xxxx}%
{Cheng \MakeLowercase{\textit{et al.}}: Deeply Explain Deep CNN}

\IEEEcompsoctitleabstractindextext{%
\begin{abstract}
\justifying   
In computer vision, some attribution methods for explaining CNNs 
attempt to study how the intermediate features affect the network prediction.
However, they usually ignore the feature hierarchies 
among the intermediate features.  
This paper introduces a hierarchical decomposition framework 
to explain CNN's decision-making process in a top-down manner.  
Specifically, we propose a gradient-based activation propagation (gAP) 
module that can decompose any intermediate CNN decision 
to its lower layers and find the supporting features.
Then we utilize the gAP module to iteratively decompose the network 
decision to the supporting evidence from different CNN layers.  
The proposed framework can generate a deep hierarchy of strongly associated 
supporting evidence for the network decision, which provides 
insight into the decision-making process.
Moreover, gAP is effort-free for understanding CNN-based models 
without network architecture modification and extra training process. 
Experiments show the effectiveness of the proposed method.
The code and interactive demo website will be made publicly available.
\end{abstract}
		
}

\maketitle
\IEEEdisplaynontitleabstractindextext
\IEEEpeerreviewmaketitle
	
\IEEEraisesectionheading{\section{Introduction}\label{sec:Introduction}}
	
\IEEEPARstart{D}{eep} convolutional neural networks (CNN) 
have made significant improvements on various computer vision tasks, 
such as image recognition \cite{simonyan2014very,he2016deep,huang2017densely}, 
object detection \cite{girshick2014rich,girshick2015fast,ren2015faster}, 
semantic segmentation 
\cite{long2015fully,chen2017deeplab,zhao2016pyramid,lin2016refinenet},
traffic environment analysis \cite{zhu2016traffic,hou2019learning},
and medical image understanding \cite{ronneberger2015u,litjens2017survey}. 
Despite the high performance, CNNs are usually used as black boxes as 
their internal decision process is unclear.
Moreover, plenty of recent research 
\cite{goodfellow2014explaining,kurakin2016adversarial,athalye2017synthesizing}, 
has pointed out 
that the previous successful CNN models can still be fooled by 
adversarial examples where the changes can not even be noticed by human eyes.
With the above prior, it is difficult for human beings to trust 
the good-performing yet opaque CNN models.
Therefore, the interpretability of CNNs is as crucial as their performance, 
especially in some critical applications.

\begin{figure}[t]
  \centering
  \begin{overpic}[width=\linewidth]{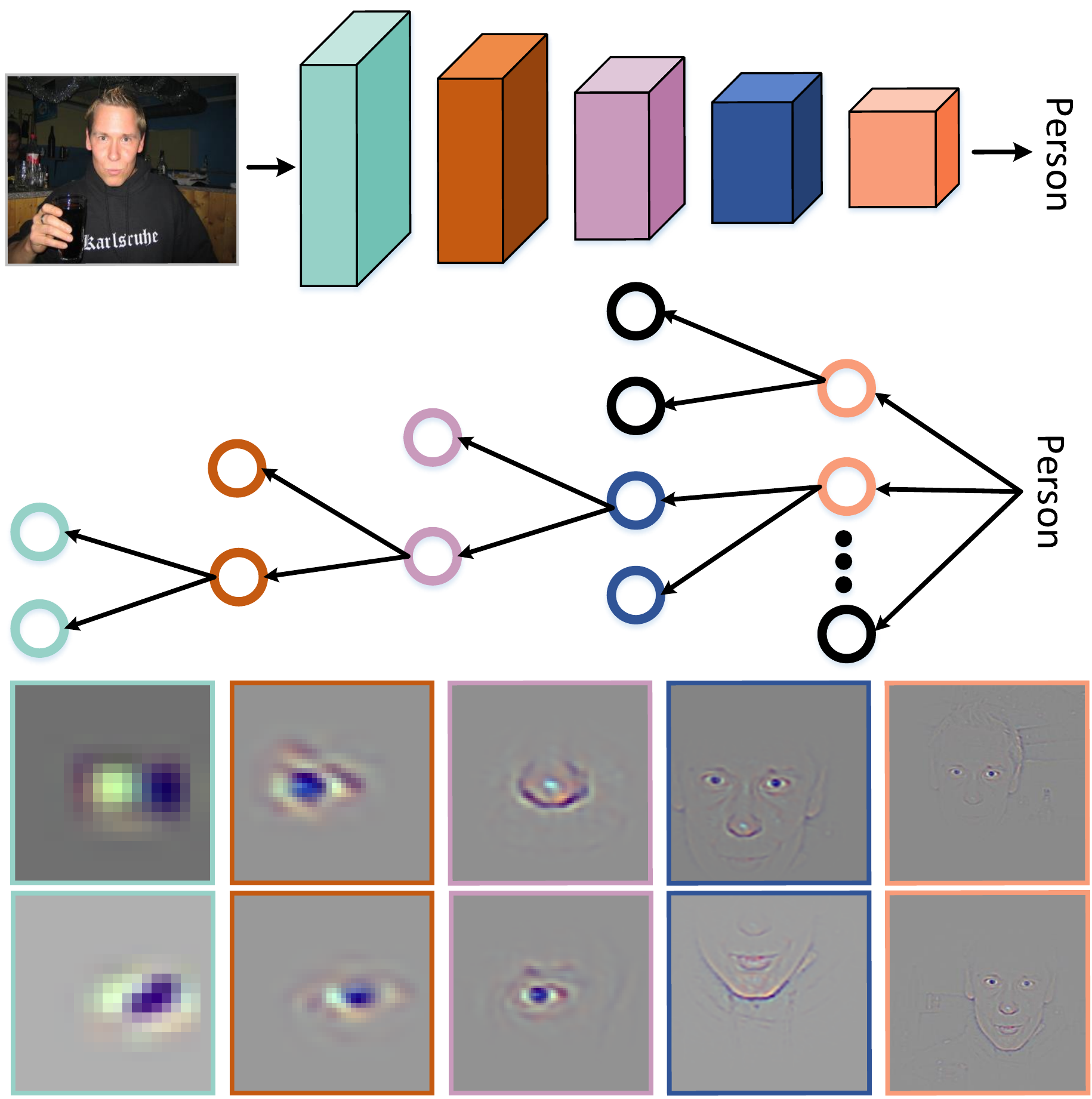}
  \end{overpic}
  \caption{Overview of our approach. 
    The middle is an evidence pyramid 
    for the network prediction, and 
    the bottom shows the important features from different 
    stages of VGG-16. 
    The colored circles represent the features.
    We detect interactions among the features and show how they 
    are combined at different hierarchies in the decision-making process. 
  }\label{fig:motivation}
\end{figure}

A fully interpretable convolutional neural network 
is a long-standing holy grail for deep learning researchers.
To this end, researchers have proposed a wide range of techniques.
Feature attribution (or saliency) methods 
\cite{sundararajan2017axiomatic,shrikumar2017learning,simonyan2013deep}
provide a powerful tool for interpretability.
They attribute an output prediction of CNN to the input image, 
where the generated saliency map can tell us which pixels 
are important to the prediction.
Such ability helps humans to understand how the input affects the prediction.
Another set of feature attribution methods
\cite{dhamdhere2018important,leino2018influence} 
measures the importance of intermediate features towards a prediction.
They further select important features and study their impact on the prediction.
Apart from generating the feature importance, 
the relationships among intermediate features \cite{olah2020zoom} 
are also important to understand the predictions but receive little attention.

CNNs have demonstrated a strong ability to gradually abstract image contents
and generate features at different semantic levels,
\eg blobs/edges, textures, and object parts
\cite{zeiler2014visualizing}.
While discovering important features can provide a rich set of evidence for
the output prediction,
isolated evidence is less convincing and informative than 
the evidence chain \cite{giannelli1982chain} 
or the evidence pyramid \cite{murad2016new}.  
According to the feature integration theory developed by 
Treisman \etal \cite{treisman1980feature},
the human brain first extracts basic features and then utilizes attention 
to combine individual features to perceive the object.
Ideally, we would expect a hierarchical evidence tree as demonstrated in
\figref{fig:motivation},
which attributes a CNN decision to multiple key features, 
and each of them can be recursively attributed to more basic features.
By associating intermediate features like 
`head', `face', `eye', `nose', and `edge' in this example, 
a group of strongly associating evidence corresponding to 
the network's inner state are emerging,
reviewing real-world facts of the human perception decision.

There are two major challenges for existing feature attribution methods 
to achieve the hierarchical decomposition.
Firstly, directly decomposing millions of feature responses 
in all channels and all spatial locations is both computationally infeasible 
and cognitively overloading for humans.
Meanwhile, feature attribution methods such as \cite{dhamdhere2018important,leino2018influence} 
are quite time-consuming because they need 
to repeat the backpropagation process many times.
Secondly, some attribution methods \cite{zhou2016learning,selvaraju2017grad} 
generate an attention map for the whole layer, 
rather than a group of attention maps for each feature channel.
The channel-wise attention maps are crucial for the iterative decomposition process 
as they can indicate the most important neuron in a feature channel to be decomposed. 
To alleviate these issues, we propose an efficient gradient-based 
Activation Propagation (gAP) module, 
which decomposes a feature response at any CNN location to its lower layer.
As the gAP module generates an activation map for each feature channel,
we can easily select a few mostly activated feature channels 
as crucial evidence,
obtaining human-scale explanations.
For each of those selected feature channels,
the CNN feature at the most activated spatial position can be 
iteratively decomposed. 
By avoiding decomposing features at too many spatial locations,
we can further reduce the number of potential visualizations to the human scale.

The proposed decomposition framework can effectively generate 
hierarchical explanations (see \figref{fig:motivation}), which 
builds relationships among crucial intermediate features.
We have conducted extensive experiments on several aspects, 
including a sanity check of the gAP module and 
understanding the network decisions. 
Experiments show the effectiveness of our framework to explain 
network decisions.
In summary, we make two major contributions: 
\begin{itemize} 
  \item We propose an efficient gradient-based Activation Propagation (gAP) module, 
  which decomposes the network decision and intermediate features 
  to find their key supporting evidence from previous layers.

  \item We propose a hierarchical decomposition framework, 
  which builds relationships among important intermediate features, 
  enabling hierarchical explanations with human-scale supporting evidence.

\end{itemize}

\section{Related Work}

The interpretability of CNNs has been actively studied,
with major progress in three main areas, 
including feature attribution, feature visualization, 
and knowledge distillation to explainable models.

\newcommand{\addFig}[1]{\includegraphics[width=0.41\linewidth]{#1}}
\newcommand{\addTexB}[1]{\textcolor{gray}{#1}}

\begin{figure} [t]
  \centering
  \subfigure[Feature Visualization \cite{springenberg2014striving}]{
    \includegraphics[width=0.470\linewidth]{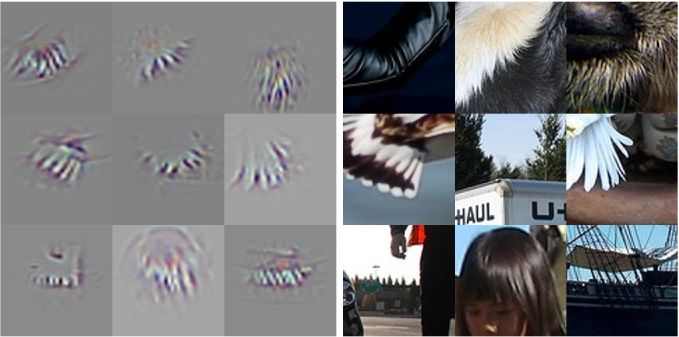}
  } 
  \subfigure[Class Activation Map \cite{zhou2016learning}]{
    \includegraphics[width=0.470\linewidth]{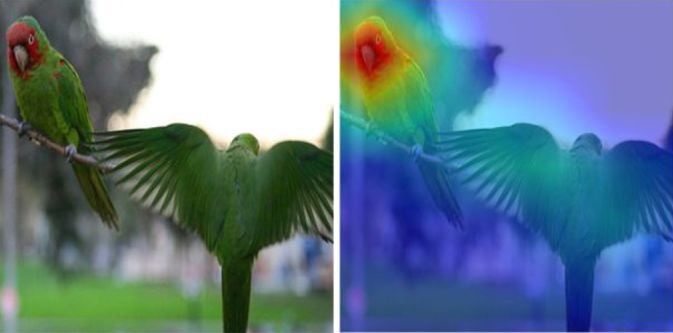}
  }\vskip 9pt 
  \subfigure[High-level Decomposition \cite{zhou2018interpretable}]{
    \begin{overpic}[width=0.96\linewidth]{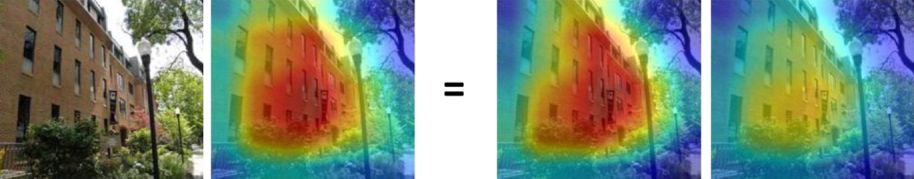} 
      \put(7,20.5){\addTexB{\footnotesize building\_facade (45\%)}}
      \put(53,20.5){\addTexB{\footnotesize balcony (9.65\%)}}
      \put(77,20.5){\addTexB{\footnotesize window (7.04\%)}}
    \end{overpic} 
  }\caption{
    Illustration of different kinds of interpretative methods.
  }\label{fig:comparison}
\end{figure}

\subsection{Feature Attribution}
Feature attribution methods typically generate a saliency map
to locate the input locations important to the output.
We classify them into three categories: 
backpropagation-based methods, perturbation-based methods, 
and activation-based methods.

\myPara{Backpropagation-based methods.}
In the early days, Sung \etal \cite{sung1998ranking} learn to 
rank the importance of input for the backpropagation networks 
by several tools such as sensitivity analysis.
Baehrens \etal \cite{baehrens2010explain} identify the feature importance 
for a particular instance by computing the gradients of the decision function.
Simonyan \etal \cite{simonyan2013deep} backpropagate the gradients 
of the output prediction \wrt the input image and generate a saliency map 
that indicates the importance of each pixel in the image. 
Guided Backpropagation \cite{springenberg2014striving} and 
Deconvnet \cite{zeiler2014visualizing} utilize different 
backpropagation logics through ReLU, 
where they both zero out the negative gradients.
Sundararajan \etal \cite{sundararajan2017axiomatic} consider 
the saturation and thresholding problem. 
They compute the saliency map by accumulating the gradients 
along a path from the base image to the input image.
Another set of methods, such as LRP \cite{bach2015pixel}, 
DeepTayor \cite{montavon2017explaining}, 
RectGrad \cite{kim2019saliency}, DeepLift \cite{shrikumar2017learning}, 
FullGrad \cite{srinivas2019full}, PatternAttribution \cite{kindermans2018learning}, 
and Excitation Backprop \cite{zhang2016top}, 
utilize different top-down relevance propagation rules. 
Yang \etal \cite{yang2020learning} attempt to learn the propagation rule 
automatically for attribution map generation.
SmoothGrad \cite{smilkov2017smoothgrad} sharpen 
the gradient-based salience maps to reduce visual noise.
Zintgraf \etal \cite{zintgraf2017visualizing} not only identify the important regions 
supporting the network decision but also identify the regions 
against the decision.
Moreover, some methods \cite{dhamdhere2018important,leino2018influence} 
measure the importance of the hidden unit to the prediction 
based on the backpropagation.
These methods can find out the most important features from different layers 
of deep networks. 
Kim \etal \cite{kim2018interpretability} study the high-level concepts 
instead of low-level features for interpreting the internal state 
of the neural network.
They utilize the directional derivatives to quantify 
the importance of high-level concepts 
to a classification result.

\myPara{Perturbation-based methods.}
These methods perturb the input to observe the output changes.
Zeiler \etal \cite{zeiler2014visualizing} occlude the input image 
by sliding a gray square and use the change of the output 
as the importance. 
Petsiuk \etal \cite{petsiuk2018rise} randomly sampled a masked region.   
Ribeiro \etal \cite{ribeiro2016should} utilize the super-pixel 
to select occluded image regions.
They learn a local linear model to compute the contribution of 
each super-pixel.
Besides, the recent methods 
\cite{fong2017interpretable,fong2019understanding,dabkowski2017real} 
learn a perturbation map, where the map applied to the input image 
can maximumly affect the prediction.
Fong \etal \cite{fong2019understanding} also apply the input 
attribution method to study the salient channels of deep networks.

\myPara{Activation-based methods.}
These methods \cite{zhou2016learning,selvaraju2017grad,chattopadhay2018grad} 
generate a coarse class activation map by linearly combining 
the feature channels from the convolutional layer.
The class activation map is upsampled to the size of the input image 
and provides image-level evidence that is important 
for the network prediction, as demonstrated in \figref{fig:comparison} (b). 
Zhou \etal \cite{zhou2016learning} propose Class Activation Mapping (CAM).
They need a specific network with the global average pooling layer 
to generate class activation maps. 
Later, Grad-CAM \cite{selvaraju2017grad} and Grad-CAM++
\cite{chattopadhay2018grad} generalize the CAM method to other tasks 
by utilizing the task-specific gradients as weights. 
Unlike Grad-CAM, Score-CAM \cite{wang2020score} utilize the forward passing 
score on the target class to obtain the weight for each activation.
Recently, Zhou \etal \cite{zhou2018interpretable} attempt to 
decompose the network decision into several semantic components 
and study each component's contribution. 
As shown in \figref{fig:comparison} (c), 
the class activation map is decomposed into several semantic components. 
\\ 

The aforementioned attribution methods mostly focus on generating 
saliency/activation maps to study how the input affects the output prediction.
Although some attribution methods can measure the importance of 
intermediate features to the output prediction, 
they usually neglect to study the relationships among different 
intermediate features.
As pointed by Olah \etal \cite{olah2020zoom}, 
the relationships among different intermediate features 
are also important to interpret a prediction.
We decompose not only the network decision but also the intermediate features 
to find their supporting evidence from previous layers,
explaining how these associated intermediate features affect each other.
While LRP \cite{bach2015pixel} method propagates feature importance 
to intermediate features,
the feature importance for different channels is coupled in the 
back-propagation process.
This method generates simple explanations for the entire network behavior 
rather than hierarchical explanations.

\subsection{Feature Visualization}
Visualizing the CNN features of the intermediate layers can 
provide insight into what these layers learn. 
For the first layer of the CNN, 
we can directly project its three-channel weights into the image space.
To visualize the features from higher layers, 
researchers have proposed many alternative approaches.
Among them, Erhan \etal \cite{erhan2009visualizing} and 
Simonyan \etal \cite{simonyan2013deep} utilize 
the gradient ascent algorithm to find the optimal stimuli 
in the image space that maximizes the neuron activations.
Other methods 
\cite{zeiler2014visualizing,springenberg2014striving,zhou2014object} 
identify the image patches from the dataset that maximize the 
neuron activation of the CNN layers, 
as shown in \figref{fig:comparison} (a).
Guided Backpropagation \cite{springenberg2014striving} and 
Deconvnet \cite{zeiler2014visualizing} also utilize
the top-down gradients to discover the patterns 
that the intermediate layers learn.
Using the natural image prior, feature inversion methods 
\cite{mordvintsev2015inceptionism,yosinski2015understanding,
mahendran2015understanding,dosovitskiy2016inverting,olah2017feature} 
learn an image to reconstruct the neuron activation.
Furthermore, the recent methods 
\cite{bau2017network,fong2018net2vec,bau2020understand} 
attempt to detect the concepts learned by intermediate CNN layers.
The above feature visualization methods explore what the intermediate 
features detect, 
but they do not answer how the network assembles individual features 
to make a prediction.

\subsection{Distill Knowledge to Explainable Models}
Recently, another research line has attempted to transfer the powerful ability 
of CNN to explainable models, such as the decision tree or linear model, 
to approximate the behavior of the original model.
Chen \etal \cite{chen2019explaining} distill the knowledge into 
an explainable additive model.
Ribeiro \etal \cite{ribeiro2016should} utilize a local linear model to 
approximate the original model, studying how the input affects 
any classifier's decisions.
Frosst \etal \cite{frosst2017distilling} and Liu \etal \cite{liu2018improving} 
distill the learned knowledge of CNN into the decision tree. 
These methods only bridge between the network decision and the input.
They cannot help the user understand how the internal features 
of CNNs affect the network decision and each other.
Our hierarchical decomposition is also an approximation to the original model.
Unlike the above methods, our hierarchical decomposition not only 
highlights the important features for the network decision but builds 
the relationships among the feature channels from different layers. 
From our method, we can obtain the states of the internal features 
and how the internal features affect each other and the network decision.

\subsection{Intrinsic Interpretable Models}
Except for the post-hoc interpretability analysis for a trained CNN, 
some researchers have attempted to explore inherently interpretable models.
Chen \etal \cite{chen2019looks} proposed a deep network architecture 
called prototypical part network.
The network has a transparent reasoning process that first computes the similarity scores 
between the image patches and the learned prototypes.
Then the network makes predictions based on a weighted sum of the similarity scores.
Concept bottleneck models \cite{koh2020concept,kumar2009attribute,lampert2009learning} 
are also inherently interpretable.
Unlike those post-hoc methods \cite{bau2017network,fong2018net2vec} that 
utilize human-specific concepts to generate explanations, 
they directly predict a set of human-specific concepts 
at training time and then use these concepts to make predictions, where 
the reasoning process is interpretable.
Some recent intrinsic interpretable models \cite{koh2020concept,chen2019looks} 
utilize VGG \cite{simonyan2014very} or ResNet \cite{he2016deep} 
to extract high-level features firstly and perform the reasoning process 
on the high-level features.
Our method is complementary to these CNN-based 
intrinsic interpretable models because one can use the hierarchical decomposition to provide 
more hierarchical evidence from the feature extractor if needed.

\section{Methodology}

\subsection{Gradient-based Activation Propagation}
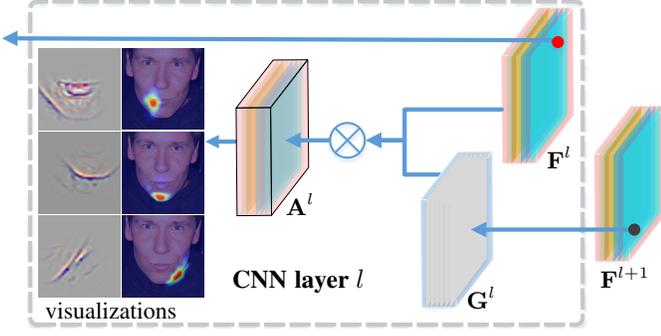
\begin{figure}[t]

  \centering
  \begin{overpic}[width=\linewidth]{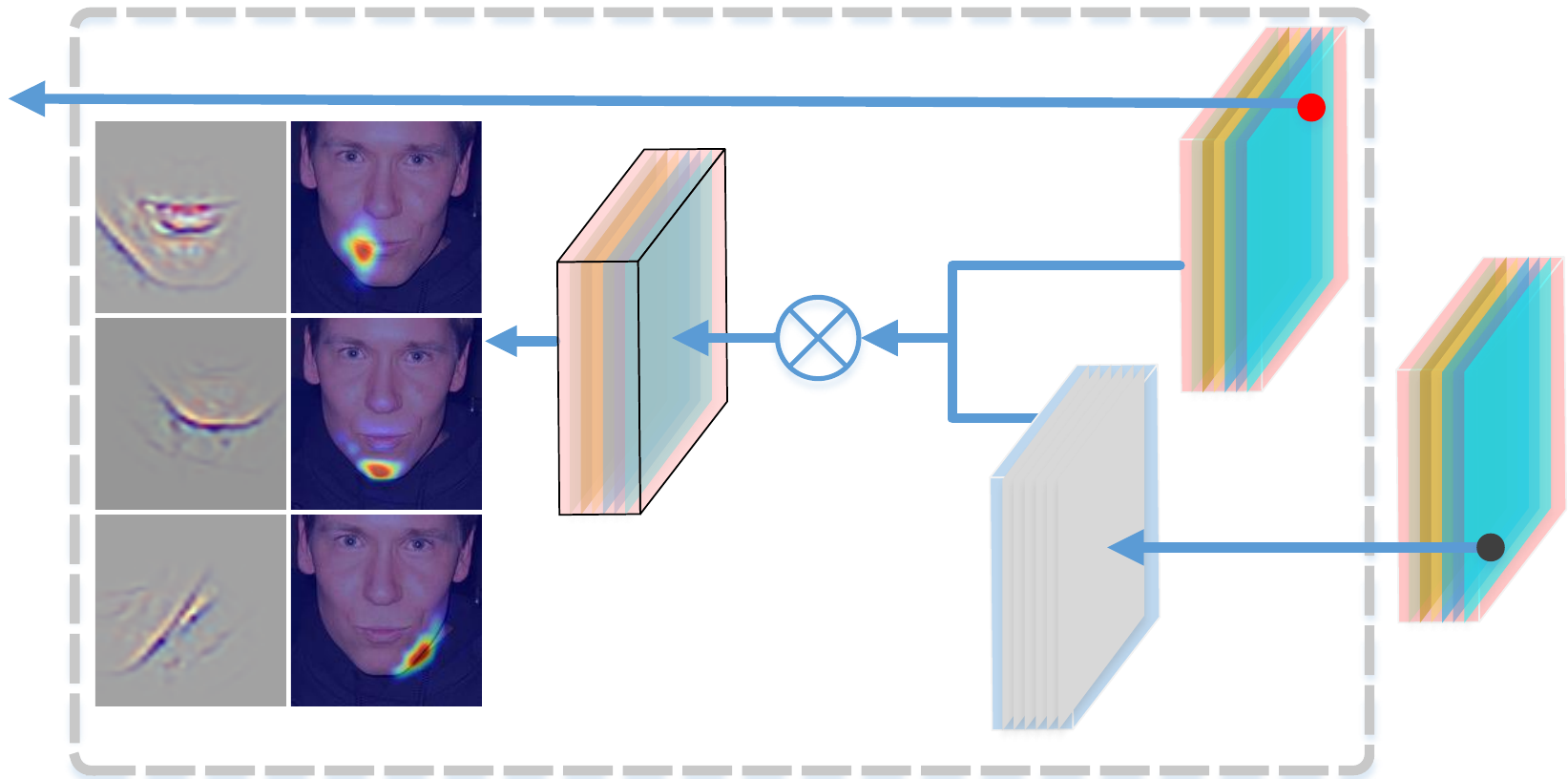}
    \put(90,6) {\small $\Feature^{l+1}$}
    \put(82,24) {\small $\Feature^l$}
    \put(70,3)  {\small $\Gradient^l$}
    \put(43,17) {\small $\Activation^l$}
    \put(7,1.5) {\small visualizations}
    \put(35,6) {\textbf{CNN layer} $l$}
  \end{overpic}
  \caption{Our gradient-based activation propagation (gAP) 
    method explains a decision of interest
    $\Feature_{k'}^{l+1}(x_{k'}^{l+1}, y_{k'}^{l+1}) \in \mathbb{R}$,  
    \ie the CNN feature illustrated by the black dot,  
    by localizing the most related neuron activations 
    in its previous CNN layer.
  }
  \label{fig:gAP}
\end{figure}

We begin by defining the notation for the CNN, 
as illustrated in \figref{fig:gAP}.
In the $l^{th}$ CNN layer,
the \textbf{features} $\Feature^l$,
partial \textbf{gradients} $\Gradient^l$,
and corresponding neuron \textbf{activations} $\Activation^l$ 
are 3D tensors with the same size, \ie 
$\Gradient^l, \Activation^l, \Feature^l \in 
\mathbb{R}^{K^l \times H^l \times W^l}$,
where $K^l$ is the number of channels and 
$H^l \times W^l$ is the spatial size in the CNN layer $l$.
To find supporting evidence for the final CNN decision 
or any intermediate feature response, 
we propose a gradient-based activation propagation (gAP) method.
Using the gAP module, we can understand 
\emph{a decision of interest at a CNN layer by localizing the most 
related evidence in its previous layer}.

As shown in \figref{fig:gAP}, 
we decompose a CNN feature (\ie a decision of interest)
$\Feature^{l+1}_{k', x, y}$ 
at the convolutional layer $l+1$, 
channel $k'$, and spatial position $(x, y)$, 
to find the supporting evidence in its previous convolutional layer $l$.
In this work, we are interested in understanding the strong feature 
response $\Feature^{l+1}_{k', x, y}$ that has the largest contribution 
to the decision among the feature channel.
In typical CNNs, a certain feature at layer $l+1$ is computed as a linear 
combination of features from its previous layer $l$ and a ReLU. 
For the strong feature $\Feature^{l+1}_{k', x, y}$, 
we have
\begin{equation} \label{eqn:linear}
  \Feature^{l+1}_{k', x, y} = 
  \text{ReLU}(\bf{w}^{l} \cdot \Feature^{l}) =
  \bf{w}^{l} \cdot \Feature^{l},
\end{equation}
where $\bf{w}_k^{l}$ is the linear weight for combining $k^{th}$ 
feature channel of $\Feature^{l}$.
To obtain the weight, we first use backpropagation 
to compute the partial gradient map $\Gradient^l_k$ 
of the feature $\Feature^{l+1}_{k', x, y}$ 
\wrt the feature map $\Feature^l_k$ by
\begin{equation} \label{eqn:grad}
\bf{w}_{k}^{l} = \Gradient_k^l = \underbrace{ 
\chfrac{\partial{\Feature^{l+1}_{k', x, y}}}
{\partial{\Feature^l_k}} 
}_\textup{gradients via backprop}. 
\end{equation}
The gradient map $\Gradient_k^l$ captures the 
`importance' of the feature map $\Feature_k^l$ for the decision
$\Feature^{l+1}_{k', x, y}$.

\begin{figure}[t]
  \centering
  \begin{overpic}[width=\linewidth]{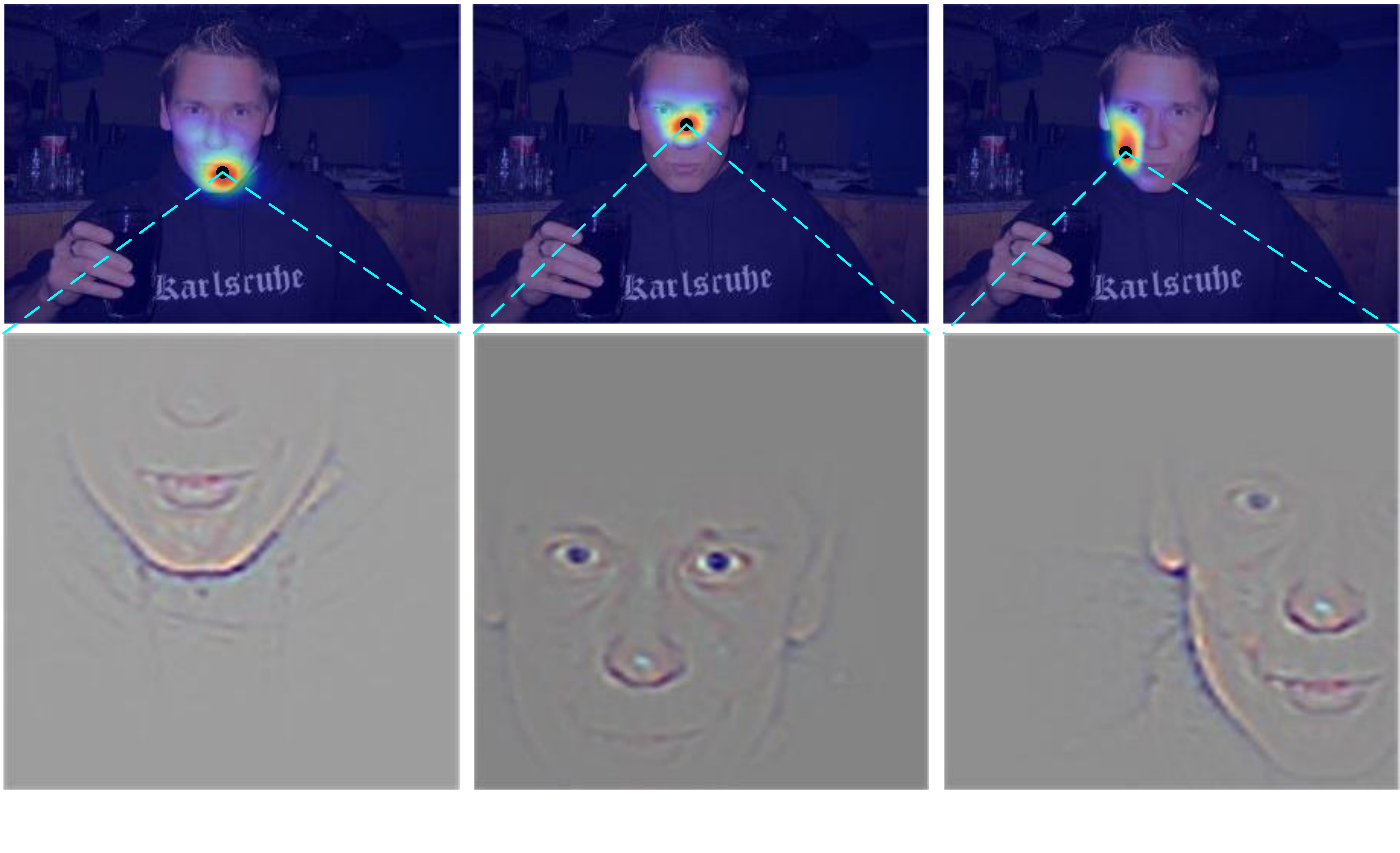}
    \put(13,0) {$131^{st}$}
    \put(45,0) {$255^{th}$}
    \put(79,0) {$452^{nd}$}
  \end{overpic}
  \caption{Example of the most significant activation maps (upper row) 
    and their corresponding visualizations (lower row) 
    for layer \conv{4}{3}, which contains $512$ channels.
    The black dot denotes the peak location in the activation map.
  }
  \label{fig:activations}
\end{figure}

We employ the gradient map $\Gradient_k^l$ to generate an activation map 
\begin{equation} \label{eqn:activations}
  \Activation^l_k = \Gradient_k^l 
  \cdot \Feature^l_k .
\end{equation}
The activation map indicates the contribution of each feature in $\Feature_k^l$ 
to the decision $\Feature^{l+1}_{k', x, y}$.
Based on its corresponding activation map, 
each channel's contribution to the decision can be computed by
\begin{equation} \label{eqn:channelImportance}
  \alpha_k^l =  \frac{1}{Z^l} \sum_x^{H^l} \sum_y^{W^l} \Activation^l_{k,x,y},
\end{equation}
where $Z^l=H^l \times W^l$ denotes the number of spatial positions 
in the activation map $\Activation^l_k$.
We can also identify the feature $\Feature^{l}_{k, \hat{x}, \hat{y}}$ 
in $k^{th}$ feature channel that contributes the most to the decision in which 
\begin{equation} \label{eqn:argmax}
  (\hat{x}, \hat{y}) = \argmax_{(x,y)} \Activation^l_{k,x,y}.
\end{equation}
Thus, for each decision, we can find the most important 
feature channel $\Feature_k^l$ according to the contribution $\alpha_k^l$ 
computed by \equref{eqn:channelImportance}.
In the most important channel, we can also identify the feature 
$\Feature^{l}_{k, \hat{x}, \hat{y}}$ that contributes most to the decision 
according to \equref{eqn:argmax}.
In the top row of \figref{fig:activations},
we show the three most important activation maps $\Activation^4_{131}$,
$\Activation^4_{255}$, and $\Activation^4_{452}$
in layer \conv{4}{3} 
for the decision from $\Feature_{277}^{5}$. 
These activation maps provide spatial channel responses to the decision, 
benefiting human understanding. 
Using Guided Backpropagation \cite{springenberg2014striving}, 
we visualize the most contributing feature by generating sharp visualizations,
which highlight the associated input.
An example is shown in the bottom row of \figref{fig:activations}.

\textbf{Discussion.} 
Our gAP module is inspired by CAM \cite{zhou2016learning} and 
Grad-CAM \cite{selvaraju2017grad},
which explain CNN decisions by class activation localization.
To explain the relation and difference to our gAP module, 
we first revisit CAM and Grad-CAM.
Smilkov \etal have proofed that Grad-CAM is a strict generalization of CAM.
Without loss of generality, we consider the same network 
discussed in \cite{zhou2016learning}.
For an image classification CNN, the CNN features $\Feature^L$ 
of the last convolutional layer are spatially pooled 
using the global average pooling layer to obtain feature vectors.
The network performs a \textbf{linear combination} of the feature vectors 
by feeding them into a fully connected layer 
before the softmax.
Let $C$ denote the number of classes.
The classification score before softmax $S^c$ for each class 
$c \in \{1, 2, \dots, C\}$ is
\begin{equation} \label{eqn:classificScore}
\begin{aligned}
S^c = \sum_k^{K^l} w^c_k \overbrace{\frac{1}{Z^L} \sum_x^{H^l} \sum_y^{W^l}}^\textup{global average pooling}  
\Feature^L_{k,x,y}\\ = \frac{1}{Z^L} \sum_x^{H^l} \sum_y^{W^l} \sum_k^{K^l} w^c_k \Feature^L_{k,x,y},
\end{aligned}
\end{equation}
where $w_k^c$ is the weight connecting the $k^{th}$ feature map 
with the $c^{th}$ class. 
The contribution of a feature $\Feature^L_{k,x,y}$ to $S^c$ 
is $w^c_k \Feature^L_{k,x,y}$. 
CAM generates a class activation map $M^c$ by summing over all feature maps,  
\begin{equation} \label{eqn:cam}
  \mathbf{M}^c = \sum_k^{K^l}  w^c_k \Feature^L_k,
\end{equation}
where each value in $\mathbf{M}^c$ indicates the contribution of each 
spatial location to $S^c$.

For the linear function, the importance weight is also equal to the gradient. 
Thus, we can also obtain the weight by computing the back-propagating 
gradients,   
\begin{equation} \label{eqn:CamWeight}
  w^c_k = \sum_x^{H^l} \sum_y^{W^l} 
  \frac{\partial{S^c}}{\partial{\Feature^L_{k,x,y}}}. 
\end{equation}
The detailed derivations of $w^c_k$ is depicted 
in \cite{selvaraju2017grad}.
\equref{eqn:CamWeight} is also the way that Grad-CAM 
computes the weight $w^c_k$.
A little difference is that Grad-CAM multiplies $w^c_k$ by 
a proportionality constant, \ie
\begin{equation} \label{eqn:GradCamWeight}
  w^c_k = \frac{1}{Z^L} \sum_x^{H^l} \sum_y^{W^l} 
  \frac{\partial{S^c}}{\partial{\Feature^L_{k,x,y}}}, 
\end{equation}
where the proportionality constant $1/Z^L$ will be normalized.

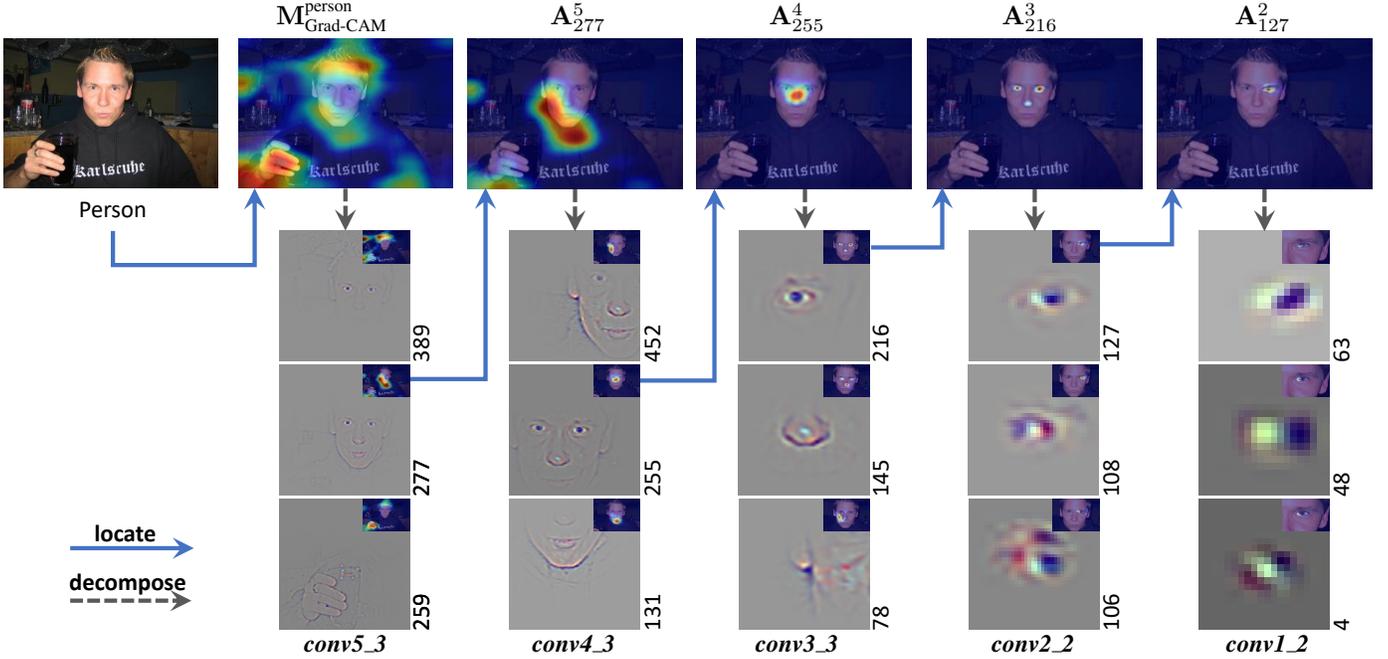
\begin{figure*}[t]
  \centering
  \begin{overpic}[width=\linewidth]{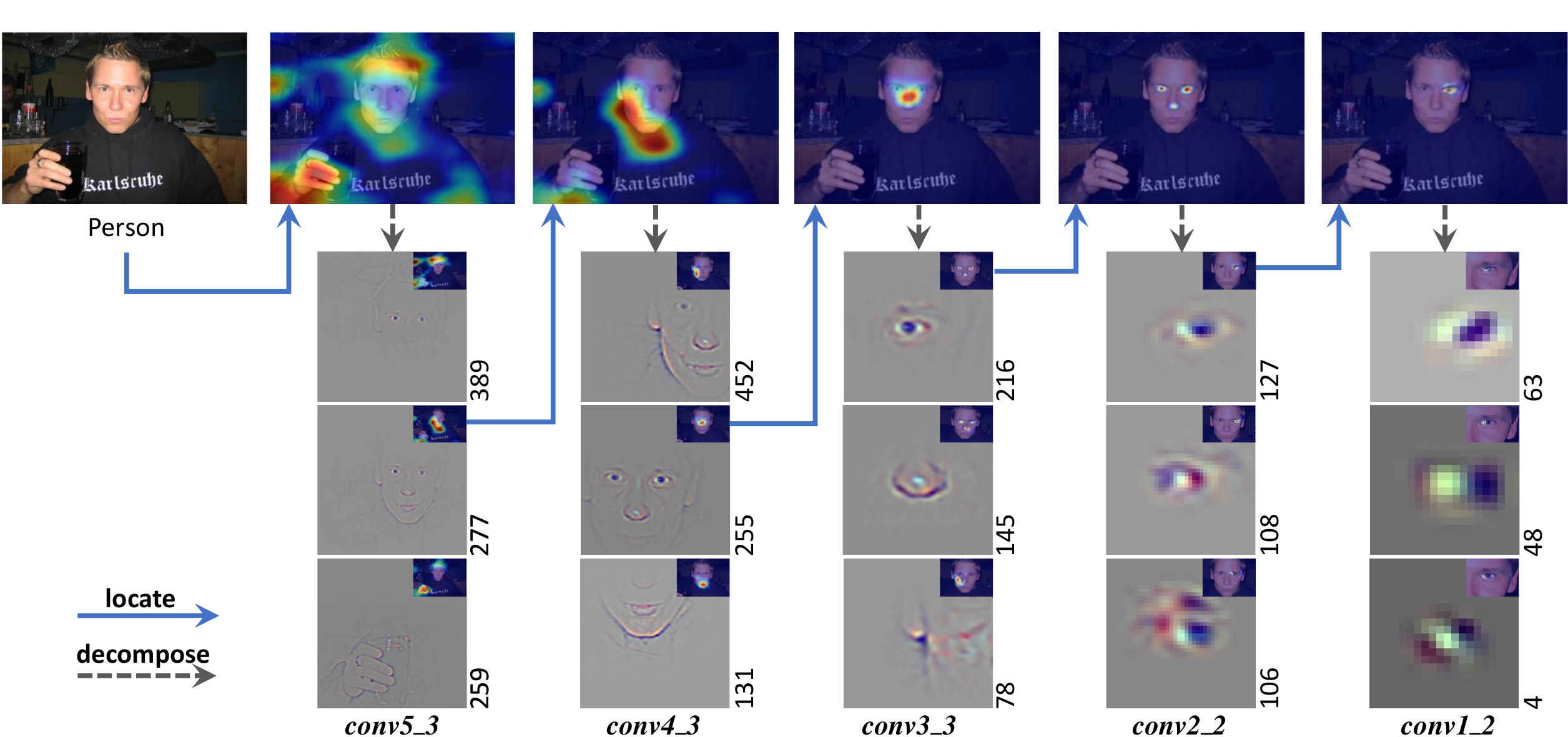}
    \put(20,46) {$\mathbf{M}_\textup{Grad-CAM}^\textup{person}$}
    \put(40,46) {$\Activation_{277}^{5}$}
    \put(56,46) {$\Activation_{255}^{4}$}
    \put(73,46) {$\Activation_{216}^{3}$}
    \put(90,46) {$\Activation_{127}^{2}$}
  \end{overpic}
  \caption{Illustration of our hierarchical decomposition process. 
    The number for each visualization denotes the channel id of 
    VGG-16 \cite{simonyan2014very}.
    At each stage, we decompose one of the top-3 most important 
    feature channels to the lower layer.
    Follow the blue line, we zoom in an activation map for the decision.
    The gray dash line represents the decomposition of the feature response
    corresponding to the maximal activation.
    Additionally, we also make the decomposition process interactive. 
    At each stage, the user can select any decision and decompose it. 
  }\label{fig:pipeline}
\end{figure*}

Considering the scores $\{S^c\}$ as CNN features with $C$ channels 
and spatial size $1\times 1$,
\ie $\Feature_{c}^{L+1}(1,1) = S^c$,
we can plug \equref{eqn:grad} into \equref{eqn:GradCamWeight} and get 
\begin{equation} \label{eqn:gAPWeights}
\begin{aligned} 
  w^c_k = \frac{1}{Z^L}\sum_x^{H^l} \sum_y^{W^l} \Gradient_{k,x,y}^L.
\end{aligned}
\end{equation}
Due to the global average pooling layer, the gradient of each element 
in $\Feature^L_k$ is the same, \ie 
$\forall_{x,y}, \Gradient_{k,x,y}^L = w^c_k$.
The class activation map $\mathbf{M}^c$ 
can be written as 
\begin{equation} \label{eqn:gradcam}
\begin{aligned}
  \mathbf{M}^c 
  = \sum_k^{K^l} \Gradient^L_{k} \cdot \Feature^L_{k} = \sum_k^{K^l} \Activation^L_{k}.  \\
\end{aligned}
\end{equation}
\equref{eqn:gradcam} suggests that the activation map $\mathbf{M}^c$ 
for Grad-CAM can be generated by simply 
adding the activations maps $\Activation^L_{k}$ from our gAP.

The differences between gAP and Grad-CAM/CAM are:
\begin{itemize}
  \item Grad-CAM/CAM combine all activation maps to generate 
  a single class activation map  $\mathbf{M}^c$, 
  which highlights important regions supporting the prediction.
  Our gAP method explains a decision of interest 
  by generating a group of activation maps $\{ \Activation_k \}$.
  Each activation map corresponds to a feature channel, 
  which is crucial for our iterative decomposition process.
  \item  Grad-CAM/CAM generate class activation maps 
  from the last convolutional layer to explain the prediction. 
  Our gAP generalizes this idea and iteratively decomposes a decision of 
  any CNN layer to its lower layer.
\end{itemize}
While the above derivations apply to adjacent layers, 
we empirically find that satisfactory decomposition results can also 
be obtained when applying the gAP module between two layers 
from different stages of CNN (see \secref{sec:sanity}).
In the following, we will describe how we build hierarchical explanations 
for the network decisions.

\subsection{Hierarchical Decomposition}
\figref{fig:pipeline} demonstrates an example of our hierarchical decomposition 
process. 
First, we decompose the network decision to the last convolutional layer 
and find the top few most crucial supporting features.
Then, we decompose each of the supporting features to their previous layer 
and iteratively repeat the decomposition process until the bottom layer.
As mentioned in \secref{sec:Introduction}, the key challenge is that 
naively building the hierarchical decomposition will generate
too many visualizations, 
which will be a cognitive burden for humans.
Even if we only decompose a single maximal contributed feature in each channel
(see also \equref{eqn:argmax}), 
directly decomposing all channels in VGG-16 will generate 
$512^3 \times 512^3 \times 256^3 \times 128^2 \times 64^2 \approx 2.5 \times 10^{22}$ visualizations.

To obtain human-scale visualizations, 
we propose two strategies to reduce the number of visualizations.
Firstly, we only decompose the top few most important features 
at each layer.
Experiment (see \secref{sec:sanity}) has verified that a small subset of feature channels 
in a layer accounts for the majority of the contributions to a decision.
Thus, we select the top few most important channels.
We simplify the top-down decision decomposition process 
by utilizing the last convolutional layer of each stage.
Current popular CNNs \cite{simonyan2014very,he2016deep} 
usually reduce the spatial size of feature maps after each stage,
where a stage is composed of a set of convolution layers 
with the same output resolution.
Each stage learns different patterns,
such as blobs/edges, textures, and object parts  
\cite{springenberg2014striving,zeiler2014visualizing}.
Experiments verify that when using the gAP module between two layers 
from two consecutive stages, 
we can obtain visually meaningful decomposition results 
(see \figref{fig:pipeline}).
By these two strategies, we can largely reduce the number of visualizations 
to obtain human-scale explanations.

An example of the VGG-16 classification network is shown in 
\figref{fig:pipeline}.
We select \conv{1}{2}, \conv{2}{2}, \conv{3}{3}, \conv{4}{3}, \conv{5}{3}
and index these layers as $\{1, 2, \dots, L\}$, where $L = 5$.
The network output before softmax could be considered as the $6^{th}$ 
CNN layer, 
with features $\Feature^{6}\in \mathbb{R}^{C \times 1 \times 1}$.
The decomposition process starts from the CNN decision $\Feature^{6}_c(1,1)$, 
where $c$ corresponds to the `person' class. 
Using gAP, we first decompose the CNN decision $\Feature^{6}_c(1,1)$ 
to $5^{th}$ layer.
The decomposition generates a set of activation maps 
$\{ \Activation^L \}$ at $5^{th}$ layer for $\Feature^{6}_c(1,1)$. 
We use \equref{eqn:channelImportance} to select the top $N$ (\eg $N$=3) 
important activation maps,
\ie $\Activation^{5}_{389}$, $\Activation^{5}_{277}$, 
and $\Activation^{5}_{259}$.
We continue to decompose the decisions from $\Feature^{5}_{389}$, 
$\Feature^{5}_{277}$, $\Feature^{5}_{259}$, 
and find the top $N$ most important activation maps at $4^{th}$ layer 
for them, respectively.
However, directly decomposing the feature map is not easy. 
Because not all of the features in a feature map contribute to decision 
(see the activation maps in the top row of \figref{fig:activations}).
We select the most representative feature that contributes most to a decision 
and decompose this feature.  
We utilize \equref{eqn:argmax} to find the feature $\Feature_{k,\hat{x},\hat{y}}^l$ 
corresponding to the maximum activation.
Then we decompose it to layer $l-1$ using gAP.
This hierarchical decomposition process recursively runs until 
we decompose the CNN decision to the lowest layer.

The number of visualizations $N$ is a flexible parameter,
which controls how many top response feature channels will be selected 
during each decomposition.
To make human cognition easier, $N$ is set to 3 in \figref{fig:pipeline}.
Moreover, we make the hierarchical decomposition interactive, 
so that the users can choose the features to be decomposed, 
easily accessing the information they need.
We also provide a video about the interactive demo, 
shown in supplementary materials.
In \figref{fig:pipeline}, we can see that the features detected 
in high-level layers can be decomposed to different parts detected 
in low-level layers.
The hierarchical decomposition process tracks important features 
and recursively explains the evidence using evidence from lower layers.
For instance, the classification results of `person' 
have been decomposed to `face' and `hand' evidence.
The `face' evidence is then decomposed to `eye', `nose', and `lower jaw'.
This process continues until we reach the lowest layer, 
which usually detects edge and blob features.

\myPara{Difference to layer-wise attribution methods.}
Some attribution methods, such as LRP \cite{bach2015pixel}, 
hierarchically propagate importance to the input 
in a layer-wise manner.
They generate a single saliency map that indicates 
the importance of each pixel in the input.
Unlike them, our method decouples the importance propagation chain 
and produces a rich hierarchy of activation maps 
and corresponding visualizations.
To explain a person image, 
our method finds \textbf{a group of} evidence, 
\eg activation maps for `face', `hand', \etc
Each evidence associated with their own supporting evidence,
\eg `face' has supporting activation maps for `eye', `nose', \etc
Our method provides informational details of 
the internal features and their relations.

\section{Experiments}
In this section, we first conduct experiments to verify 
the correctness and efficiency of the decision decomposition.
Then, we use the hierarchical decomposition process 
to analyze network characteristics and explain network decisions. 
We conduct experiments on two popular datasets, ImageNet 
\cite{russakovsky2015imagenet} and PASCAL VOC \cite{everingham2015pascal}. 
On the PASCAL VOC dataset, the augmented training set 
containing  10582 training images is used to fine-tune different 
classification networks.
All the experiments are tested on a single RTX 2080Ti GPU.

\subsection{Sanity Check for gAP} \label{sec:sanity}
\myPara{The effectiveness of gAP.}
We have shown that the gradient-based Activation Propagation (gAP) module
helps to decompose the network decision hierarchically for the CNN-based 
models.
During the decomposition process,
what matters most is the accuracy of the channel contributions
calculated by the gAP module.
Thus, we first examine the accuracy of the channel contributions
to the decision of interest.
Following \cite{dhamdhere2018important,zhang2019interpret,bau2020understand},
we take the decision score drop,  
when removing a feature channel at a time, as the ground truth of the channel's contribution.
 Specifically, given an input image $I$, 
let $f^{l+1}$ be a decision score at the $l+1^{th}$ layer.
$\hat{f}^{l+1}$ denotes the decision score when setting 
the $k^{th}$ feature channel in the $l^{th}$ layer to the average activation.
The score drop $\hat{\alpha}_k^l = f^{l+1} - \hat{f}^{l+1}$ denotes the 
 $k^{th}$ channel's ground truth contribution to this decision.

\begin{table}[t]
  \centering
  \small
  \renewcommand{\arraystretch}{1.05}
  \setlength\tabcolsep{1.2mm}
  \caption{The Pearson correlation coefficient (PCC) of different settings.  
  $\rightarrow$ denotes the decomposition.
  \textbf{S5-S1} denotes the last convolutional layer of 5 different stages 
  in VGG-16 \cite{simonyan2014very}.
  \textbf{AA}: Average Activation. \textbf{MA}: Maximum Activation.
  \textbf{AG}: Average Gradient. \textbf{MG}: Maximum Gradient. 
  \textbf{T}: target category.  
  Average activation achieves the best result.
  } \label{tab:pcc}
  \begin{tabular}{cccccc} \toprule[1pt]
    ImageNet & T $\rightarrow$ S5 & S5 $\rightarrow$ S4  & S4 $\rightarrow$ S3 
     & S3 $\rightarrow$ S2 & S2 $\rightarrow$ S1  \\ 
    \midrule[0.8pt]
    AA & 0.985 & 0.959 & 0.933 & 0.898 & 0.895 \\  
    MA & 0.897 & 0.912 & 0.894 & 0.864 & 0.890   \\  
    AG & 0.623 & 0.421 & 0.497 & 0.545 & 0.472  \\  
    MG & 0.454 & 0.456 & 0.567 & 0.594 & 0.606   \\   
    \midrule[0.8pt]   
    \midrule[0.8pt] 
    VOC & T $\rightarrow$ S5 & S5 $\rightarrow$ S4  & S4 $\rightarrow$ S3 
     & S3 $\rightarrow$ S2 & S2 $\rightarrow$ S1  \\ 
    \midrule[0.8pt]
    AA &  0.987  &  0.961  & 0.932 & 0.899 & 0.893     \\  
    MA &  0.917  &  0.913  & 0.892 & 0.856 & 0.897     \\  
    AG &  0.702  &  0.492  & 0.525 & 0.564 & 0.480     \\  
    MG &  0.575  &  0.525  & 0.536 & 0.583 & 0.669     \\      
     \bottomrule[1pt]
  \end{tabular}
\end{table}

The Pearson Correlation Coefficient (PCC) metric \cite{benesty2009pearson} 
is utilized to measure the linear correlations between ground truth contribution
$\hat{\alpha}^l \in \mathbb{R}^{K^l}$ and the contribution 
$\alpha^l \in \mathbb{R}^{K^l}$
estimated by \equref{eqn:channelImportance}.
When the PCC value equals 1, there are linear correlations 
between the two variables 
(0 denotes no linear correlations, -1 denotes total 
negative linear correlations). 
The PCC metric is computed by 
\begin{equation} 
\rho = \frac{\mathbb{E}\left[(\alpha^l - \mu_{\alpha^l})(\hat{\alpha}^l - \mu_{\hat{\alpha}^l})\right]}{\sigma_{\alpha^l} \cdot \sigma_{\hat{\alpha}^l}} ,
\end{equation}
where $\mu$ and $\sigma$ denote the mean and variance, respectively.

As shown in \tabref{tab:pcc}, we study several strategies of calculating 
the contribution of a feature channel to the decision of interest.
It can be seen that the contribution computed by averaging activations 
(\ie \equref{eqn:channelImportance})
obtains the highest PCC value with the ground truth. 
For all stages in VGG-16, there are strong linear correlations between
the computed contributions and the ground truth.
This high correlation verifies the effectiveness of the gAP module.

Taking the computational efficiency into account, 
it is rather a time-consuming style of measuring channel contributions
by calculating the score drop when removing feature channels iteratively in a layer
\cite{dhamdhere2018important,zhang2019interpret,bau2020understand}.
In comparison, only one backpropagation process is needed
when gAP calculates the channel contributions to a decision.
On a VGG-16 backbone, calculating the ground truth channel contributions
of an image takes about 10s, while the gAP module only takes about 50ms, 
nearly 200x faster.
With the efficiency advantages of the gAP module, 
our hierarchical decomposition process can immediately yield 
detailed explanations of a network decision.

\begin{figure}[t]
  \centering
  \begin{overpic}[width=\linewidth]{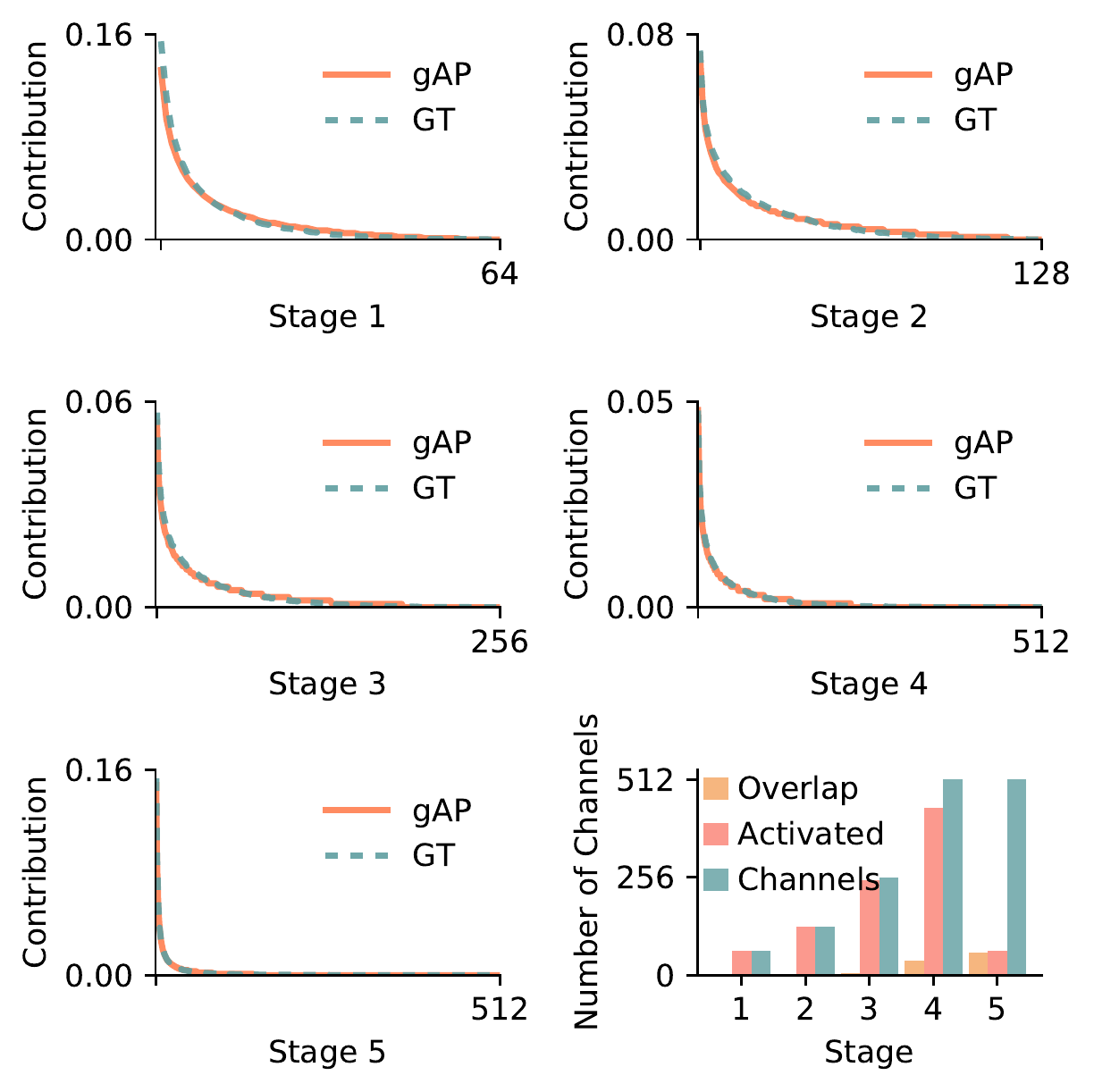}
  \end{overpic}
  \caption{ (a) The first five figures plot the contributions of each channel 
                in a CNN layer to the decision.
        The contribution of a channel to a decision denotes how much it affects the decision.
    gAP denotes the proposed gradient-based activation propagation method. 
    GT denotes the method that removes feature channels. 
      `Stage 1-5' denotes the last convolutional layer of each stage.
      The channel contributions are sorted in descending order.
      The contribution distribution calculated by gAP keeps almost
      the same as that of the ground truth.
      Besides, the contribution distribution in a layer is long-tailed.
      (b) The last chart plots the number of the activated channels 
      and the number of all channels at different layers.
      In high-level layers, there are many activated channels 
      with similar effects to a decision. 
  }
  \label{fig:statis}
\end{figure}

\myPara{The distribution of contributions.} 
As shown in the first five curves of \figref{fig:statis}, 
we can observe that the distribution of the channel contributions 
in a CNN layer is long-tailed.
A small number of feature channels play the most important role for 
a decision of interest.
With deeper layers of the networks, 
the proportion of the important feature channels decreases.
In high-level layers, the feature channels are usually more discriminative.
This fact is in line with the accepted notion \cite{zeiler2014visualizing}.
Besides, we also check how many feature channels at a CNN layer work 
together to determine a decision for the higher CNN layers.
We call the channel with $\alpha_k^l  > 0$ as the activated channels 
and compute the number of the activated channels in the decision 
decomposition process.
As shown in the last chart of \figref{fig:statis},
when decomposing a decision from layer \conv{2}{2} to layer \conv{1}{2},
nearly all channels in layer \conv{1}{2} are found activated.
However, for the decomposition from the final decision 
to layer \conv{5}{3}, we can see that the activated channels' number 
is much less than the number of all channels in layer \conv{5}{3}.

\begin{figure}[t]
  \centering
  \subfigure[model randomization test]{
    \begin{overpic}[width=0.575\linewidth]{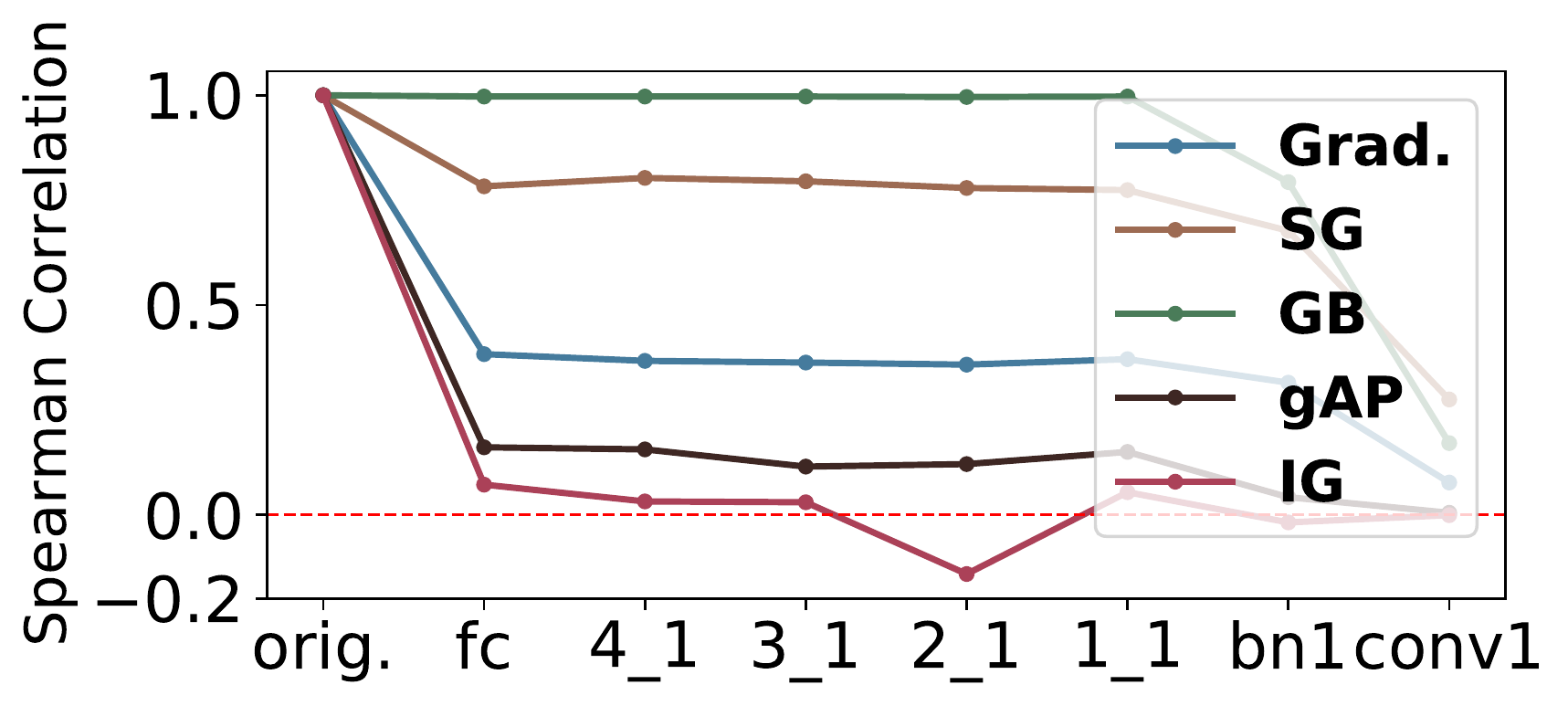}
    \end{overpic}}
  \subfigure[data randomization test]{
    \begin{overpic}[width=0.385\linewidth]{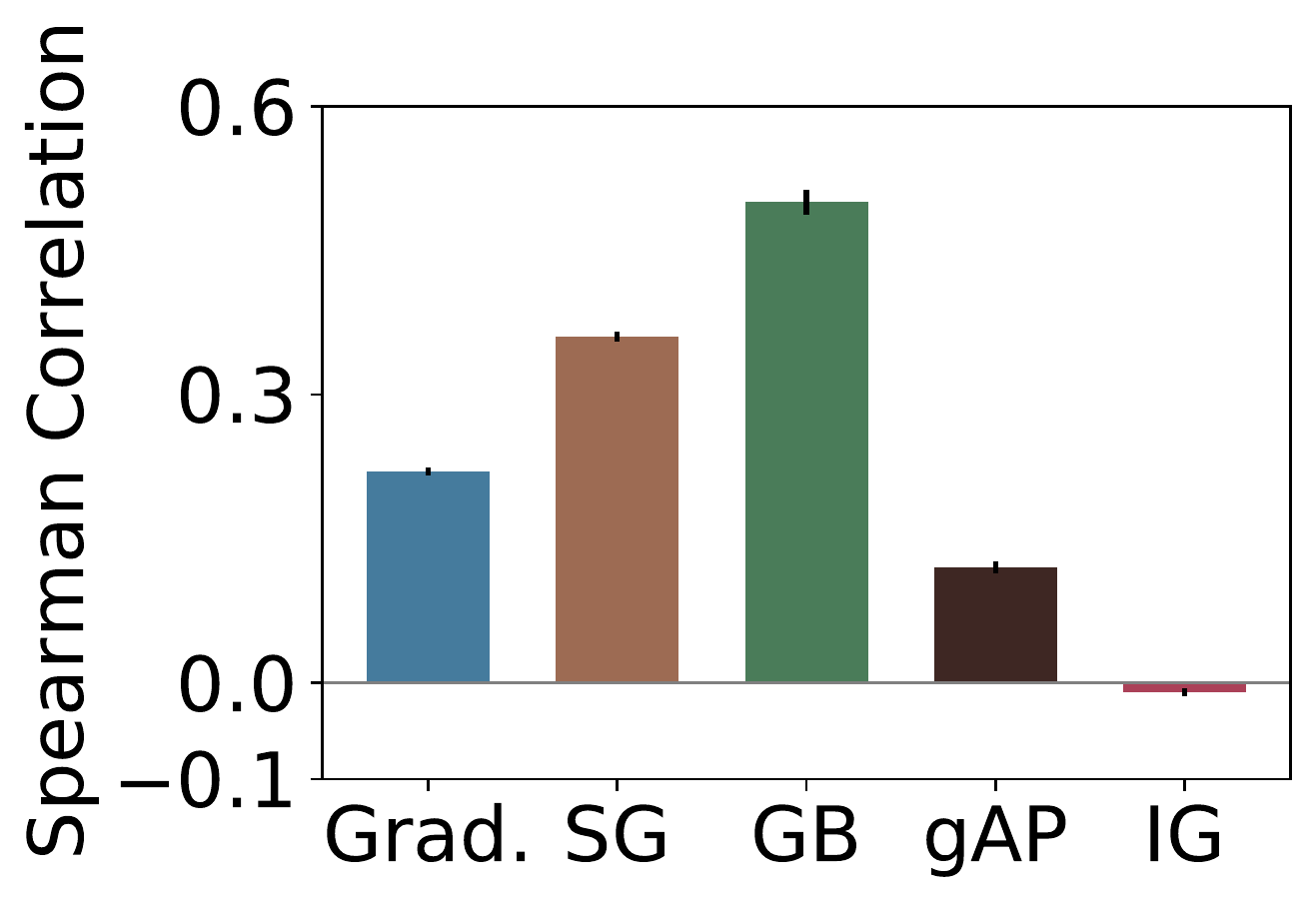}
  \end{overpic}}
  \caption{ Sanity check for different attribution methods 
    using cascading model parameter and data randomization test.  
  SG: SmoothGrad \cite{smilkov2017smoothgrad}, 
  GB: Guided Backprop \cite{springenberg2014striving}, 
  IG: Integrated Gradient \cite{sundararajan2017axiomatic}.
  The spearman rank correlation metric \cite{sedgwick2014spearman} 
  is used to measure the correlation between the attribution maps 
  of the original model and the randomizing model.
  Low correlation means the attribution method is sensitive to the 
  model parameters and the data labeling,
  and thus suitable for explaining the model decisions.
  Our gAP obtains low correlation values in these two tests.
  Best viewed with zoom in.
  }
  \label{fig:model_test}
\end{figure}

\myPara{Channel-effect overlaps.} 
Using the gAP module, we observe that the activation 
maps of some channels decomposed from the same decision 
often have strong activations in similar spatial locations.
Such spatial locations usually denote an underlying concept 
\cite{bau2017network,fong2018net2vec}, contributing to the decision.
When presenting visualizations of the hierarchical 
decomposition, we will merge these duplicate channels with similar effects 
for human better understanding.
Specifically, when decomposing a decision of interest into the 
lower layer, we will obtain activation maps corresponding 
to each channel in this layer.
We first threshold the activation maps into binary masks 
and then compute Intersection-over-Union (IoU) between them.
Then we apply the non-maximum suppression algorithm \cite{neubeck2006efficient} 
to suppress activation maps with an IoU score larger than 0.9, 
where the activation maps are sorted using the contribution scores 
by \equref{eqn:channelImportance}.
As shown in \figref{fig:statis}, we present how many activated channels 
have large overlaps with each other.
In low-level layers, the number of activated channels with large overlaps 
is very small. 
But in high-level layers, there are many activated channels 
with similar effects to a decision.

\myPara{Sanity checks for gAP.}
Adebayo \etal \cite{adebayo2018sanity} propose the model parameter 
and data randomization test for sanity check for visual attribution methods.
These two tests are used to check whether the attribution method is 
sensitive to the model parameters and the labeling of the data.
An attribution method insensitive to the model parameters and data labels is 
inadequate for debugging the model and explaining the mechanism that depends on the relationship between the instances and the labeling of the data.
To generate saliency maps from our gAP, we hierarchically 
decompose the decision until the data layer 
and sum all the gradients from each decomposition.
We do the model parameter randomization test 
on the pretrained ResNet-18 model \cite{he2016deep} and randomly 
initialize the model parameters from the top layer to the bottom layer 
in a cascading manner.
We utilize the spearman rank correlation metric to compute the difference 
between the attribution maps from the original model and the randomly 
initialized model.
Besides, we do the data randomization test by comparing the saliency maps 
from CNNs trained with true labels and permuted labels, respectively.

In \figref{fig:model_test}(a), the low spearman metric indicates 
that the attribution maps from the original model and the randomly initialized model 
differ substantially, which demonstrates that gAP is sensitive to model parameters.
In \figref{fig:model_test}(b), the low spearman metric also indicates 
gAP is sensitive to the labeling of the data.
The experimental results verify that our method can be used for debugging models.
The visual comparisons are shown in supplementary materials.

\begin{figure}[t]
  \centering
  \begin{overpic}[width=0.92\linewidth]{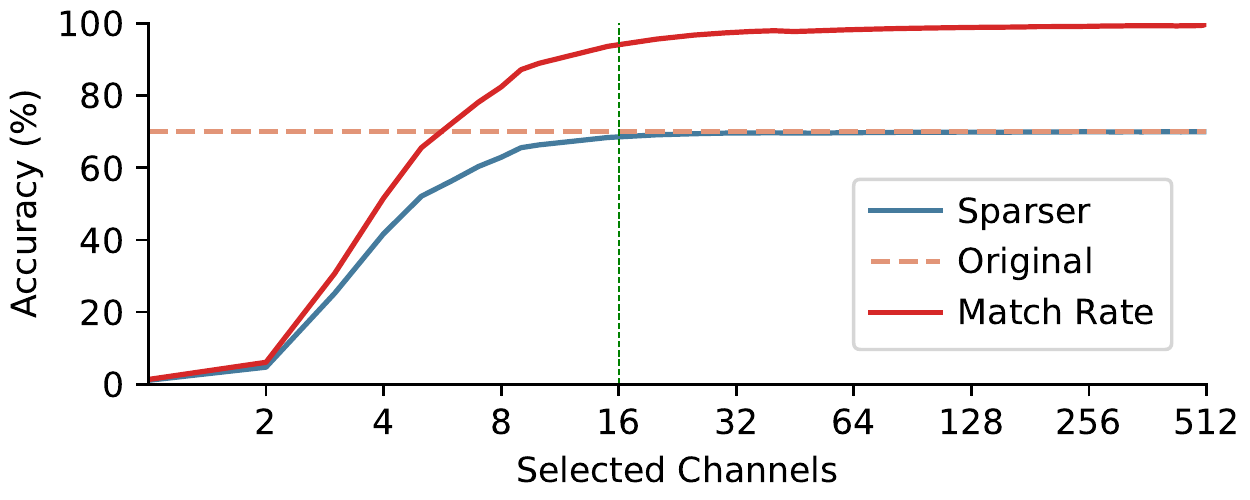}
  \end{overpic} 
  \caption{ The classification accuracy of the sparser model generated by gAP 
  and the original model. 
  The match rate denotes the prediction agreement between the sparser model 
  and the original model.
  }
  \label{fig:prune}
\end{figure}

\begin{figure}[t]
  \centering
  \begin{overpic}[width=0.98\linewidth]{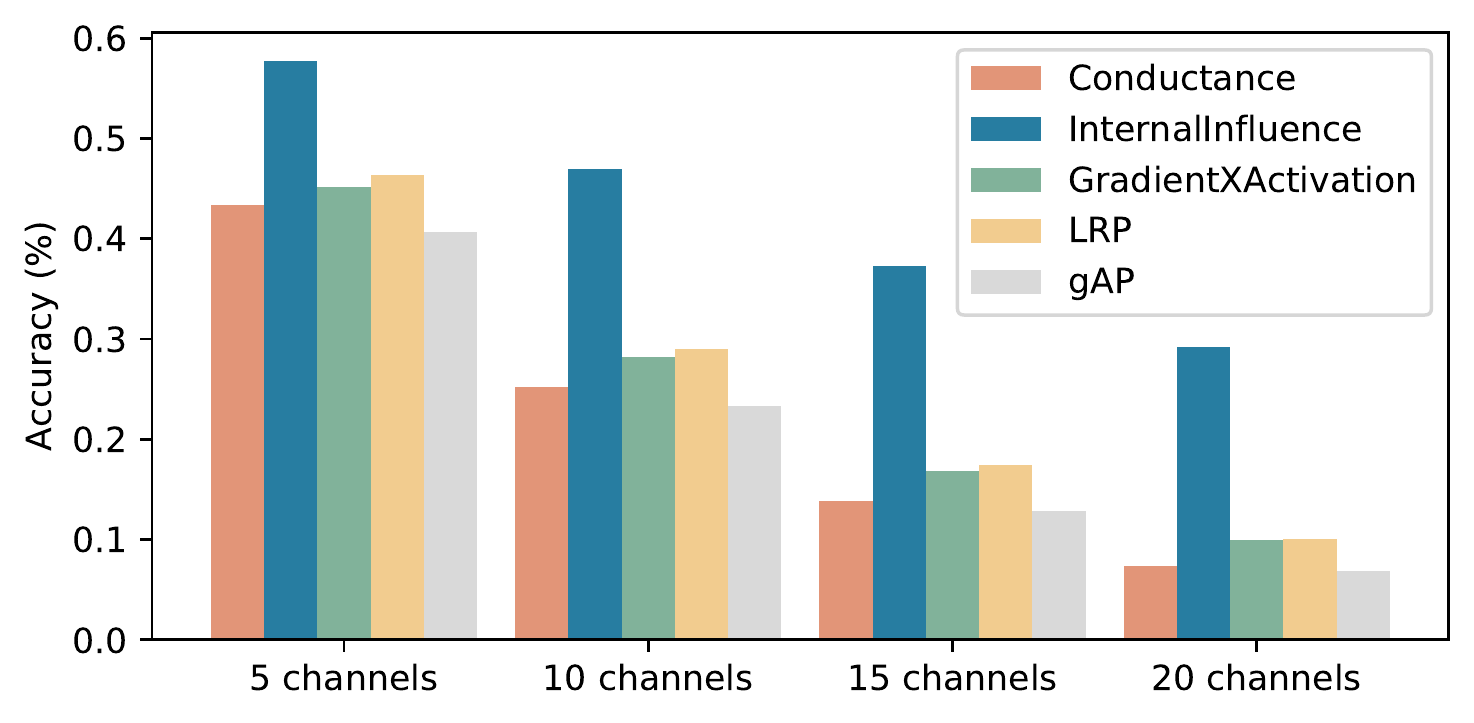}
  \end{overpic}
  \caption{ Comparisons of the classification accuracy after removing 
  the top few most important feature channels (the lower the better). 
  The y-axis is the classification accuracy on the ILSVRC validation set \cite{krizhevsky2012imagenet}. 
  The x-axis means the number of important feature channels to ablate.
  Conductance: \cite{dhamdhere2018important}, InternalInfluence: \cite{leino2018influence}, 
  LRP: \cite{bach2015pixel}. 
  }
  \label{fig:importance}
\end{figure}

\myPara{Is the top-k decomposition a good approximation to 
the original model? }
We have tested the classification accuracy of the sparser surrogate model 
generated by gAP.
Moreover, we measure the match rate by comparing the predictions 
between the sparser surrogate model and the original model.
Specifically, we decompose from the decision of the predicted category 
to the bottom layers and select the top-k important features 
in each decomposition to make predictions. 
As shown in \figref{fig:prune}, when using top-16 decomposition, 
the sparser surrogate model has a similar classification accuracy 
to the original model.
According to the match rate, when using top-16 decomposition,  
the predictions of the sparser surrogate model and the original model 
are consistent on almost all samples.
The sparser surrogate model selecting a small number of feature channels 
can make a good approximation to the original model.

\myPara{Comparison with individual-based methods.}
the individual-based methods \cite{dhamdhere2018important,leino2018influence} 
compute the importance of each channel from different layers 
to the final network decision.
Compared with individual-based methods, 
gAP can help us explore the relationships 
among different feature channels.
To directly compare with them, we propagate the importance of each selected channel 
of the top layer to the shallow layer.
We select the top-$N$ most important channels from different layers of VGG-16 
and ablate them to watch the change of the classification accuracy.
We conduct experiments on the ILSVRC validation set \cite{krizhevsky2012imagenet}.
As shown in \figref{fig:importance}, 
when removing the top few most important feature channels, 
gAP obtains lower classification accuracy 
than other individual-based methods.
We analyze that gAP only propagates the contributions of those important feature channels 
to lower layers, which reduces the interference of other feature channels.
Compared with the individual-based methods, 
gAP can not only effectively detect the important features 
but also how these features affect each other.

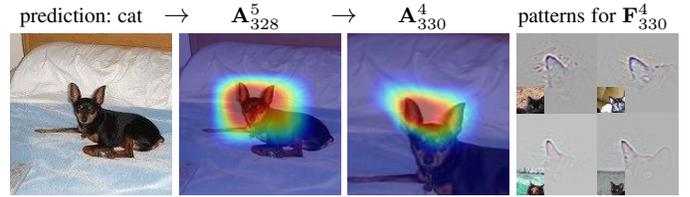
\begin{figure}[t]
  \centering
  \begin{overpic}[width=\linewidth]{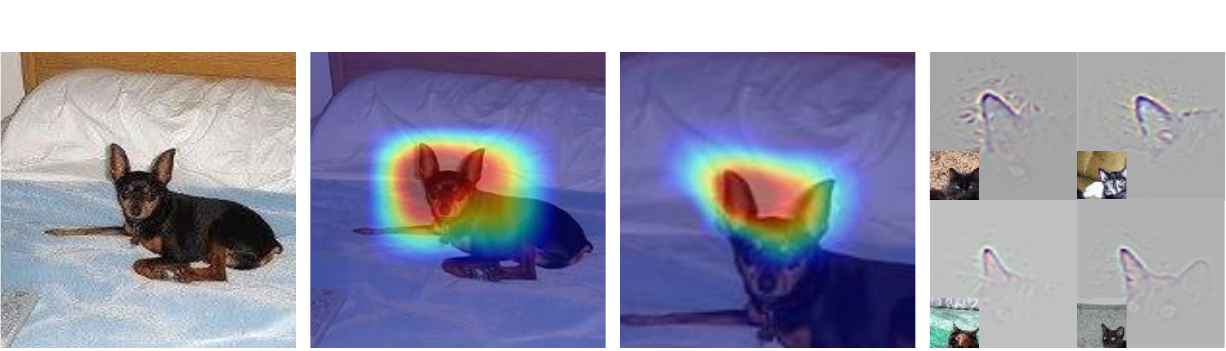}
  \put(1.5, 26.0){\footnotesize prediction: cat}
  \put(23, 26.0){$\rightarrow$}
  \put(33, 26.0){\footnotesize $\Activation_{328}^{5}$}
  \put(48, 26.0){$\rightarrow$}
  \put(58, 26.0){\footnotesize $\Activation_{330}^{4}$}
  \put(76, 26.0){\footnotesize patterns for $\Feature_{330}^{4}$ }
  \end{overpic}
  \caption{Analysis of a failure example. 
    The leftmost is the dog image that 
    is misclassified to the \emph{cat} category. 
    We decompose the network decision to layer \conv{4}{3}.
    The rightmost is the patterns that maximumly activate the $330^{th}$ channel. 
    For this example, the channels sensitive to the \emph{cat} category's attribute 
    have strong activations, causing VGG-16 to make a wrong decision.
  }\label{fig:failure}
\end{figure}

\subsection{Diagnosing CNN}

\myPara{Analyzing failure predictions of CNN.}
Previous work \cite{selvaraju2017grad} can generate class 
activation maps for the network predictions, 
highlighting the most important image regions 
supporting the network decision.
However, such an explanation is not informative enough.
The hierarchical decomposition can further provide 
a more detailed explanation for the network decision.
We decompose the network's decision iteratively 
to the low-level layers and find the most important feature channels 
at different layers.
We can see each channel's contribution to the network decision. 
Further, important channels and their corresponding activation maps 
can also be studied. 

\begin{figure}[t]
  \centering
  \subfigure[Decomposition]{
    \begin{overpic}[width=0.98\linewidth]{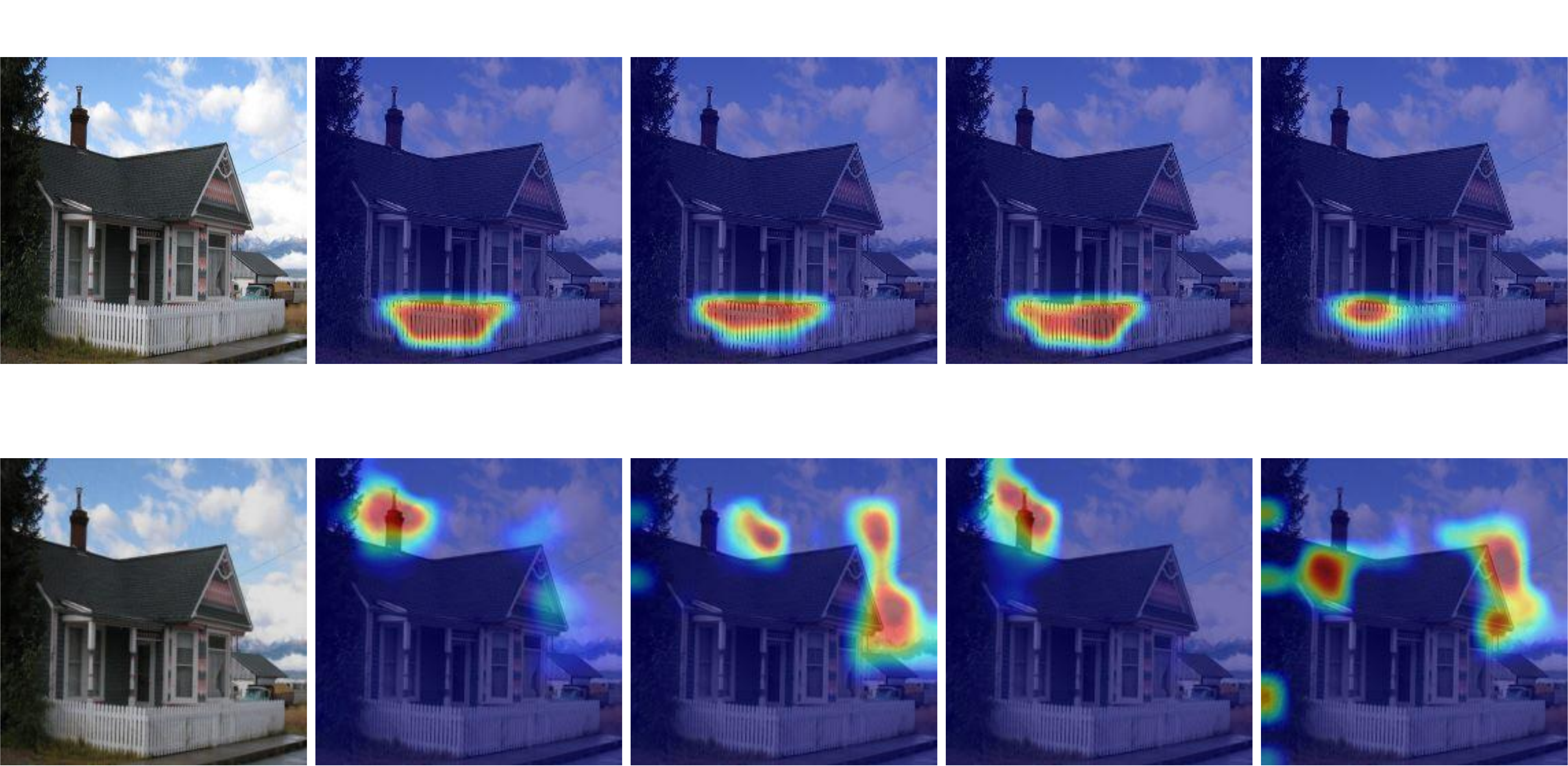}
    \put(1.5, 46){\footnotesize picket\_fence }
    \put(20, 46){$\rightarrow$ }
    \put(27, 47){\footnotesize $\Activation_{181}^{5}$ }
    \put(47, 47){\footnotesize $\Activation_{146}^{5}$ }
    \put(67, 47){\footnotesize $\Activation_{100}^{5}$ }
    \put(87, 47){\footnotesize $\Activation_{499}^{5}$ }
    \put(5, 20.5){\footnotesize church }
    \put(18, 21){$\rightarrow$ }
    \put(27, 21){\footnotesize $\Activation_{33}^{5}$ }
    \put(47, 21){\footnotesize $\Activation_{440}^{5}$ }
    \put(67, 21){\footnotesize $\Activation_{188}^{5}$ }
    \put(87, 21){\footnotesize $\Activation_{466}^{5}$ }
    \end{overpic}
  } 
  \subfigure[Peak Response]{
    \includegraphics[width=0.98\linewidth]{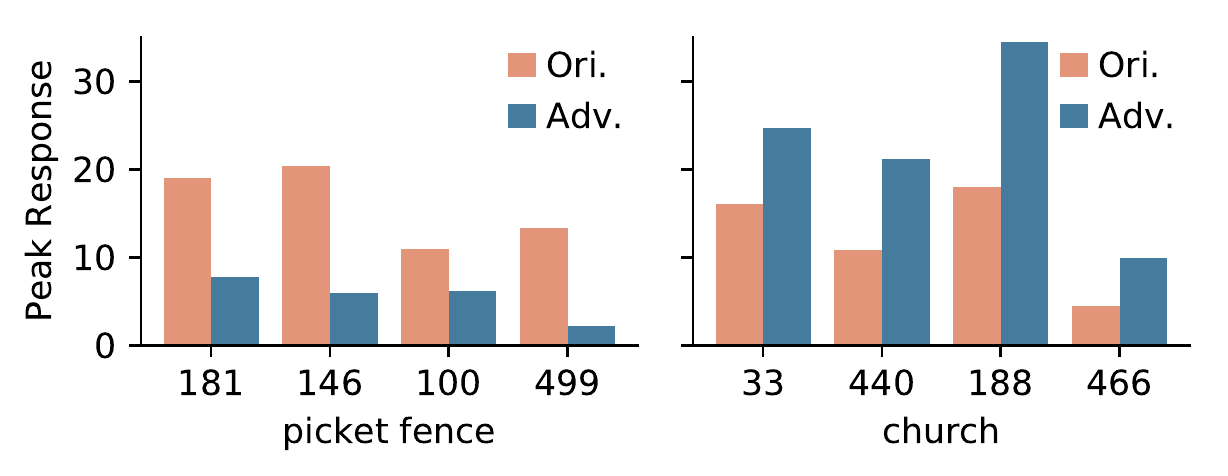}
  }  
  \subfigure[Average Peak Response]{
    \includegraphics[width=0.98\linewidth]{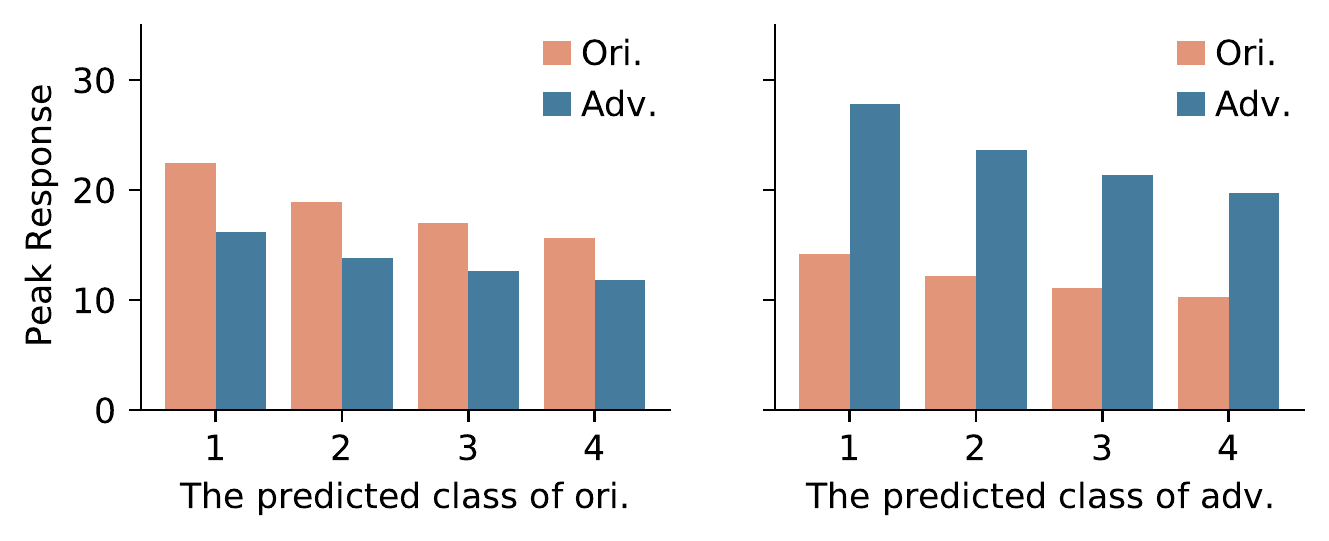}
  }  
  \caption{Example of the adversarial attacks. 
    (a) The top is the original image, and the bottom is the adversarial image. 
    $\rightarrow$ denotes the decomposition.
    (b) plots the peak feature responses of the most important channels for 
    the network decision in the original image and adversarial image.
    (c) plots the average peak feature responses of the top 4 most important channels 
    for the network decision in the original image and adversarial image 
    over the whole ILSVRC validation dataset \cite{krizhevsky2012imagenet}. 
    The peak values of the important channels for the correct category largely decrease, 
    and those for the wrong category increase by a large margin.
  }\label{fig:adv}
\end{figure}
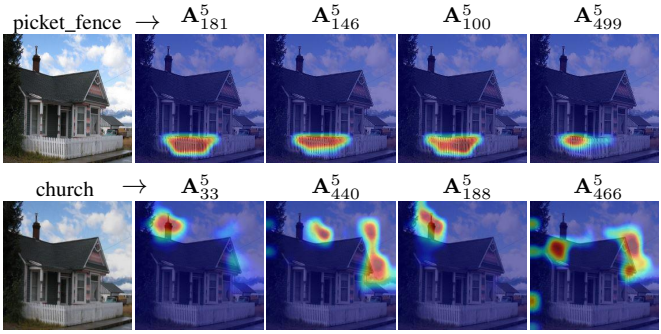
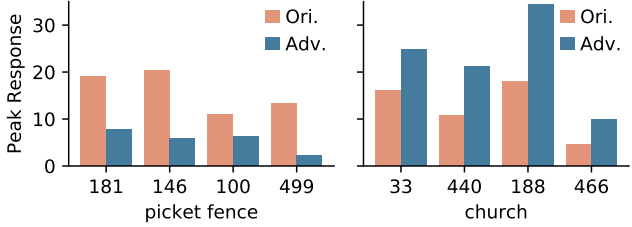
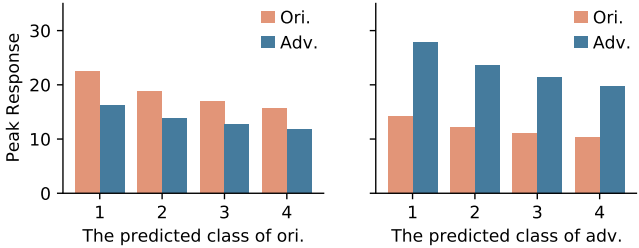

As shown in \figref{fig:failure}, we use the hierarchical decomposition 
to examine the CNN's wrong decision.
\figref{fig:failure} demonstrates a failure case. 
A dog image misclassified to the \emph{cat} category with a probability of 99\%.
We first decompose the network decision to layer \conv{5}{3} 
and find the most important channel, \ie the $328^{th}$ channel, 
with a 32.3\% contribution. 
We further present the decomposition from channel $328^{th}$ to 
layer \conv{4}{3} and find the most important channel, 
\ie the $330^{th}$ channel.
The activation map $\Activation_{330}^{4}$ has strong activations 
at the ear region.
Moreover, the patterns that maximumly activate the $330^{th}$ channel 
are the ear image patches of the \emph{cat} category.
We find the dog's ear of this example has a similar shape 
to those ear image patches of the \emph{cat} category.
We further occlude the image region of the dog's ear 
and observe that the CNN correctly predicts the \emph{dog} category 
with a probability of 65\%.
With the hierarchical decomposition, we found that CNN makes the wrong decision
because it takes the dog's ear as the cat's ear in this example.

\myPara{Analyzing adversarial attacks.}
Current CNN models are vulnerable to adversarial attacks.
When the adversarial attack algorithms add a small perturbation to 
the original images, these CNN models easily misclassify them.
To understand how the adversarial images successfully fool the CNN models, 
following \cite{bau2020understand}, 
we study the change of the feature responses for important channels.
As shown in \figref{fig:adv}(a), we present the original image (top row) 
and the adversarial image (bottom row).
The adversarial image is generated by a popular attack algorithm 
\cite{madry2017towards}.
VGG-16 classifies the original image to the \emph{picket$\_$fence} category 
(probability $92\%$) and the adversarial image to the 
\emph{church} category (probability $100\%$).
Through our decomposition from the network decision to layer \conv{5}{3}, 
we find the top few most important feature channels for the \emph{picket$\_$fence} 
and \emph{church} category, respectively.

As shown in \figref{fig:adv}(b), when comparing the adversarial 
image to the original image, we observe that the peak feature 
responses of important channels for the \emph{picket$\_$fence} category, 
\ie the $181^{st}$, $146^{th}$, $100^{th}$, $499^{th}$ channels, 
largely decrease by 11.3, 14.5, 4.7, and 11.1.
However, the peak feature responses of important channels for 
the \emph{church} category, \ie the $33^{rd}$, $440^{th}$, $188^{th}$, $466^{th}$ channels, 
largely increase by 8.7, 10.4, 16.4, and 5.5.
As shown in \figref{fig:adv}(c), we also compute the average peak responses 
of important channels on the whole ILSVRC validation dataset \cite{krizhevsky2012imagenet}.
The adversarial attack algorithms change the feature responses 
of important channels to affect the final network decision.
For the important channels, 
they reduce the correct category's feature responses and 
increase the wrong category's feature responses.

\begin{figure}[t]
  \centering
  \begin{overpic}[width=\linewidth]{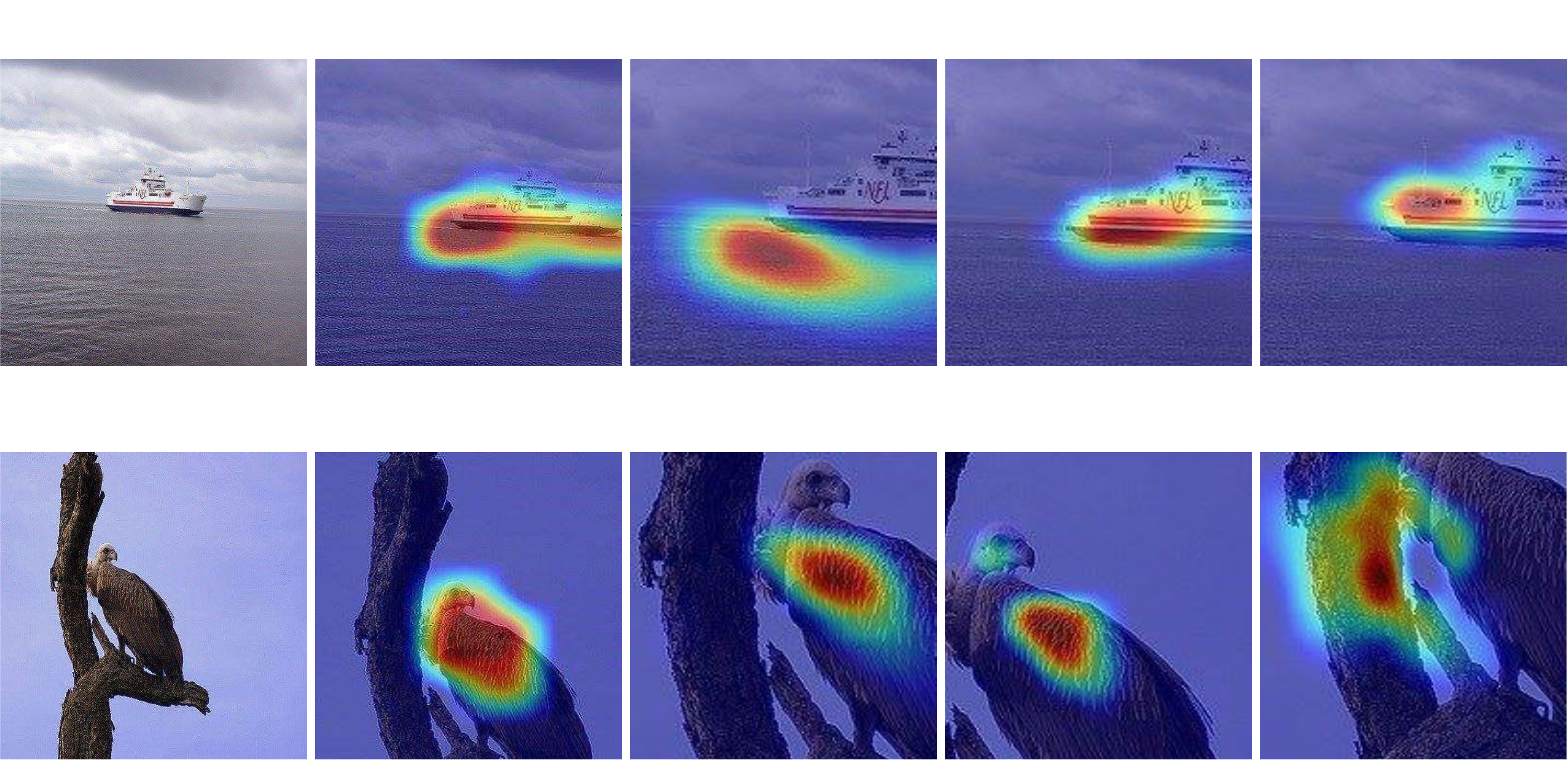}
    \put(8, 46){\footnotesize boat  }
    \put(18, 46){$\rightarrow$ }
    \put(27, 46.5){\footnotesize $\Activation_{331}^{5}$  }
    \put(40, 46){$\rightarrow$ }
    \put(47, 46.5){\footnotesize $\Activation_{169}^{4}$  }
    \put(67, 46.5){\footnotesize $\Activation_{404}^{4}$  }
    \put(86, 46.5){\footnotesize $\Activation_{115}^{4}$  }

    \put(8, 21){\footnotesize bird }
    \put(18, 21){$\rightarrow$ }
    \put(27, 21.5){\footnotesize $\Activation_{270}^{5}$ }
    \put(40, 21){$\rightarrow$ }
    \put(48, 21.5){\footnotesize $\Activation_{1}^{4}$ }
    \put(67, 21.5){\footnotesize $\Activation_{435}^{4}$ }
    \put(86, 21.5){\footnotesize $\Activation_{175}^{4}$ }
  \end{overpic}
  \caption{The context information in the activation maps. 
     $\rightarrow$ denotes the decomposition.
   }
  \label{fig:context}
\end{figure}

\begin{figure}[t]
  \centering
  \begin{overpic}[width=\linewidth]{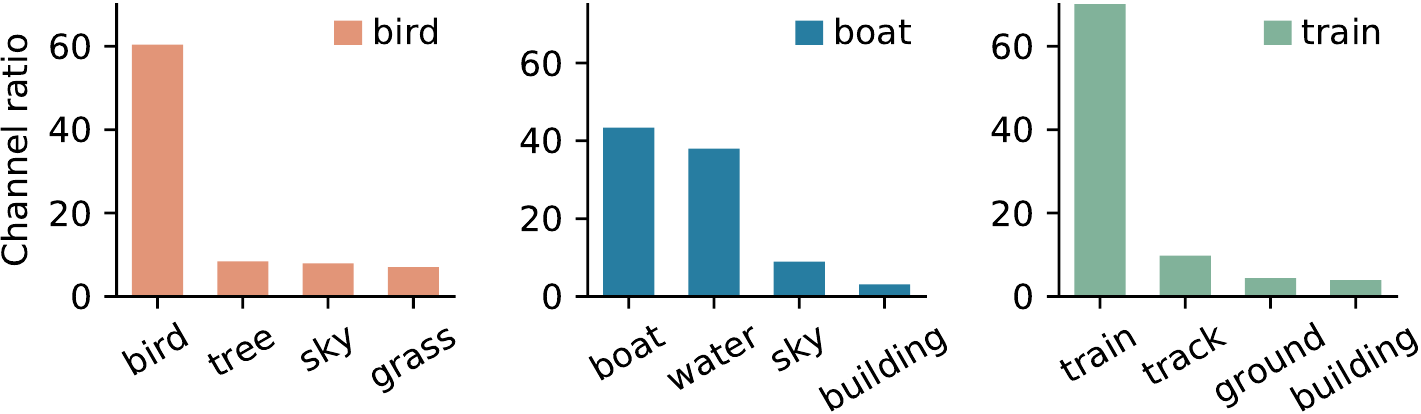}
  \end{overpic}
  \caption{Context information for each category in the PASCAL VOC dataset.
  }
  \label{fig:context_statis}
\end{figure}

\myPara{The context in activation maps. }
Context information \cite{oquab2015object,kumar2005hierarchical} 
is crucial for recognition.
A known prior is that the target category usually 
appears in a specific context.
For example, the boats usually appear in the seas or lakes,  
and the birds often stand on the tree branch.
Through our decision decomposition, we find some context 
in the activation maps to support the CNN prediction.
\figref{fig:context} shows that the $331^{st}$ channel in layer 
\conv{5}{3} has strong responses to the image's `boat' region. 
We decompose the peak point indicated by the activation map 
to layer \conv{4}{3}.
The $169^{st}$,  $404^{th}$, and $115^{th}$ channels are the top-3 
most important channels.
The most important channel is the $169^{st}$ channel, 
whose corresponding activation map locates the sea.

To quantitatively analyze the context information 
contained in the activation maps, we utilize the PASCAL-Context 
dataset \cite{mottaghi2014role} for evaluation.
We select the images with context annotations from the PASCAL VOC 
validation set \cite{everingham2015pascal}
and compute the most frequent context labels for each category.
Specifically, we perform the hierarchical decomposition to 
layer \conv{4}{3}, obtaining the activation map 
for each selected channel.
The activation map is first thresholded to a binary map.
Then we compute the IoU between the binary activation map 
and each context region.
The activation map is assigned with the label of the context region 
corresponding to the largest IoU.
In \figref{fig:context_statis}, we have shown the top few most frequent 
context labels for three categories, \ie bird, boat, and train.
These categories usually appear in a specific 
environment.
This fact suggests that the context of the objects is critical for recognition.
The context information of other categories and 
the qualitative examples are shown in the supplementary materials.

\myPara{Channel discrimination analysis.}
We utilize the hierarchical decomposition 
to explore the discriminative information of the channels 
in different layers.
Specifically, we define a discriminative degree $D$ 
to measure the discriminative information of a channel.
When performing the hierarchical decomposition process for 
the images with label $c$, 
we count the number of times $N_c$ for channel $k$ 
when its contribution to a decision ranking top-3. 
$N_c$ is summed on all images from the validation set.
Then the discriminative degree $D$ 
is computed by 
\begin{equation} 
D = \frac{\max \limits_{c} N_c}{\sum_{c=1}^C N_c},
\end{equation}
where $C$ denotes the number of categories in the dataset. 
When the channel $k$ is only decomposed from one single category, 
the discriminative degree $D$ = 1.
Besides, we can get the minimum value of $D$ when the channel 
decomposed from each category with equal times: $D = 1 / C$.

\begin{figure}[t]
  \centering
  \begin{overpic}[width=\linewidth]{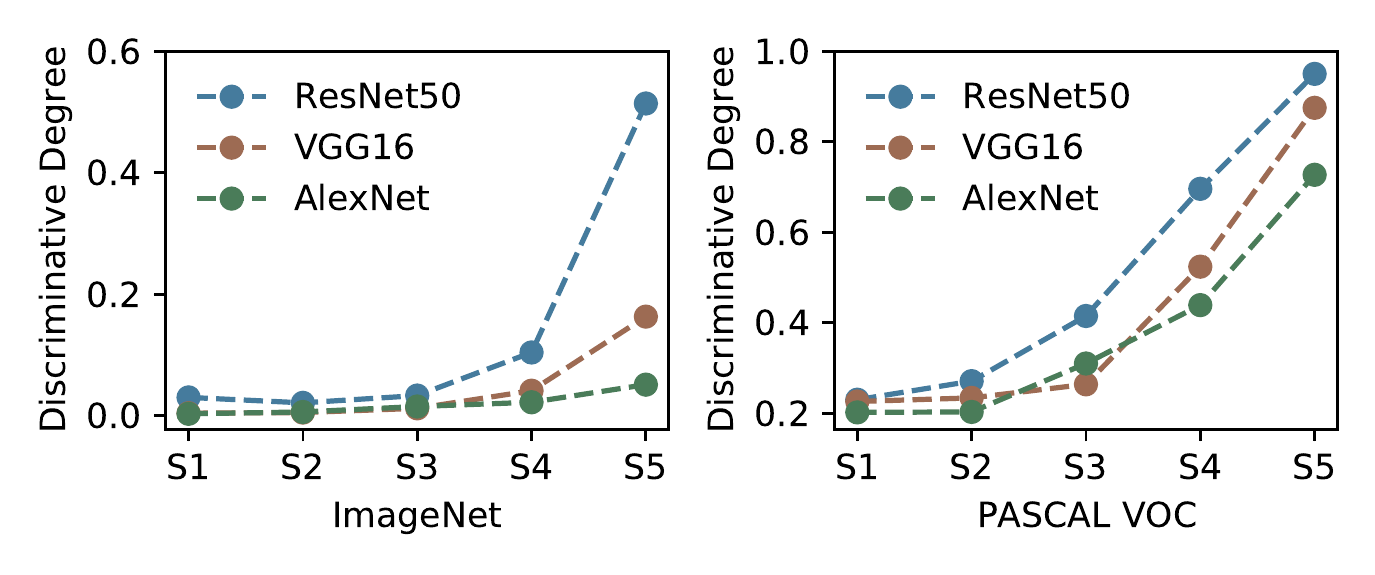}
  \end{overpic}
  \caption{ The discriminative degrees of the disentangled channels from different layers 
          of different CNNs on the validation set of PASCAL VOC \cite{everingham2015pascal} 
          and ImageNet \cite{russakovsky2015imagenet}.  
          }
  \label{fig:discrim}
\end{figure}

We apply the hierarchical decomposition to different CNNs. 
As shown in \figref{fig:discrim}, the channels' discriminative 
degrees in low-level layers are very small.
They usually have strong activations for multiple categories.
This fact indicates that the basic features detected by channels 
in low-level layers are shared among different categories, 
which lacks discriminative information for classification.
However, in high-level layers of CNNs, the channels' 
discriminative degrees are much larger than those in low-level layers.
Because the high-level layers in CNNs gradually combine basic features 
from low-level layers to form more discriminative features.
In high-level layers, different categories tend to highlight their own 
discriminative channels. 
These results provide additional evidence for the conclusion found 
by Zeiler \etal \cite{zeiler2014visualizing}.

Moreover, for the high-level layers of different CNNs, 
the discriminative degrees of the channels gradually increase with 
the growth of the network depth 
(ResNet-50 \cite{he2016deep} $>$ VGG-16 \cite{simonyan2014very} $>$ 
AlexNet \cite{krizhevsky2012imagenet}). 
Such difference suggests that the high-level layers of ResNet-50 
have a stronger discriminative ability. 
The strong discriminative ability of the channels can effectively 
reduce confusion among different categories, 
which helps ResNet-50 achieve higher classification accuracy 
than VGG-16 and AlexNet.

\section{Limitation}
The proposed hierarchical decomposition method explains 
the individual decision by selecting a set of 
strongly correlative channels from different layers of CNN.
These feature channels provide a rich hierarchy of evidence.
However, the feature channels are less confident for an unprofessional user 
to understand the network's reasoning process because not all examples 
are as easy to understand as the person image.
So in the future, we will attempt to build connections 
between the selected feature channels and human-specific concepts 
for better human understanding.
Besides, following \cite{dhamdhere2018important,bau2020understand}, 
we have removed channels individually to study their contributions.
However, as verified in \cite{fong2018net2vec,leavitt2020selectivity}, 
the representations are usually distributed among multiple channels.
We observe that the activation maps of some channels 
decomposed from the same decision often have strong activations 
in similar spatial locations.
This phenomenon suggests that multiple feature channels produce 
class responses together.
One possible solution to the flaw of removing channels individually is 
that we can first find those feature channels with similar effects 
by measuring the overlap between their corresponding activation maps.
Then we analyze these feature channels together to the network decision.
In this paper, we focus on building the evidence hierarchy.
The issue of removing individual channels will be our future work.

\section{Conclusion}

We present a novel gradient-based activation propagation (gAP) scheme 
that can decompose any CNN layer's decision to its lower layers.
Based on the gAP, the network decision can be hierarchically decomposed to 
a rich set of the evidence pyramid associated with all layers of the CNN model.
Our method allows users to delve deep into the CNN's decision-making process 
in a top-down manner.
We have experimentally verified the effectiveness of our method and 
demonstrated its ability to understand and diagnose CNN predictions.
While currently mostly focus on explaining CNN-based image classifiers,
we will study how to generalize the framework to other tasks and 
other deep learning models in the future.
The source code and interactive demo website will be made publicly available.

\bibliographystyle{IEEEtran}
\bibliography{hd}

\end{document}